\documentclass[letterpaper]{article} 
\usepackage{aaai24}  
\usepackage{times}  
\usepackage{helvet}  
\usepackage{courier}  
\usepackage[hyphens]{url}  
\usepackage{graphicx} 
\usepackage{natbib}  
\usepackage{caption} 
\usepackage{algorithm}
\usepackage{algorithmic}

\usepackage{xcolor}
\usepackage{multirow}
\usepackage{adjustbox}
\usepackage{amssymb}
\usepackage{subcaption}
\usepackage{dsfont}
\usepackage{bm}
\usepackage{amsmath}
\usepackage{pifont}
\usepackage{diagbox}
\newcommand{\ie}{\textit{i}.\textit{e}.}
\newcommand{\eg}{\textit{e}.\textit{g}.}
\usepackage{colortbl}
\usepackage{newfloat}
\usepackage{listings}
\DeclareCaptionStyle{ruled}{labelfont=normalfont,labelsep=colon,strut=off} 
\floatstyle{ruled}
\newfloat{listing}{tb}{lst}{}
\floatname{listing}{Listing}
\title{SurgicalSAM: Efficient Class Promptable Surgical Instrument Segmentation}
\author{
    Wenxi Yue\textsuperscript{\rm 1}, 
    Jing Zhang\textsuperscript{\rm 1}, 
    Kun Hu\textsuperscript{\rm 1}, 
    Yong Xia\textsuperscript{\rm 2}, 
    Jiebo Luo\textsuperscript{\rm 3}, 
    Zhiyong Wang\textsuperscript{\rm 1}
}
\affiliations{
    \textsuperscript{\rm 1}School of Computer Science, The University of Sydney\\
    \textsuperscript{\rm 2}School of Computer Science, Northwestern Polytechnical University\\
    \textsuperscript{\rm 3}Department of Computer Science, University of Rochester
 \\
    \{wenxi.yue, jing.zhang1, kun.hu, zhiyong.wang\}@sydney.edu.au, yxia@nwpu.edu.cn, jluo@cs.rochester.edu

}

\usepackage{bibentry}

\begin{document}

\maketitle

\begin{abstract}
The Segment Anything Model (SAM) is a powerful foundation model that has revolutionised image segmentation. To apply SAM to surgical instrument segmentation, a common approach is to locate precise points or boxes of instruments and then use them as prompts for SAM in a zero-shot manner. However, we observe two problems with this naive pipeline: (1) the domain gap between natural objects and surgical instruments leads to inferior generalisation of SAM; and (2) SAM relies on precise point or box locations for accurate segmentation, requiring either extensive manual guidance or a well-performing specialist detector for prompt preparation, which leads to a complex multi-stage pipeline. To address these problems, we introduce SurgicalSAM, a novel end-to-end efficient-tuning approach for SAM to effectively integrate surgical-specific information with SAM's pre-trained knowledge for improved generalisation. Specifically, we propose a lightweight \textit{prototype-based class prompt encoder} for tuning, which directly generates prompt embeddings from class prototypes and eliminates the use of explicit prompts for improved robustness and a simpler pipeline. In addition, to address the low inter-class variance among surgical instrument categories, we propose \textit{contrastive prototype learning}, further enhancing the discrimination of the class prototypes for more accurate class prompting. The results of extensive experiments on both EndoVis2018 and EndoVis2017 datasets demonstrate that SurgicalSAM achieves state-of-the-art performance while only requiring a small number of tunable parameters. The source code is available at \textit{https://github.com/wenxi-yue/SurgicalSAM}.
\end{abstract}

\section{Introduction}
Surgical instrument segmentation (SIS) is a crucial task in surgical vision, aimed at precisely delineating surgical instruments in operative scenes. It provides vital assistance to surgeons and facilitates the development of advanced computer-assisted operation systems \cite{surgery, memory, skill, skill_multitask, cmtnet_workflow, aIoT}. 
Existing deep learning methods for SIS have achieved impressive results through the design and training of specialist models featuring task-specific components. Nevertheless, these methods usually require training the complete set of model parameters (\ie, full training) using SIS datasets, resulting in inefficiency. In addition, due to the limited scale of the SIS datasets, the trained models tend to exhibit subpar generalisation performance.

\begin{figure}[!t]
\centering
\includegraphics[width=0.46\textwidth]{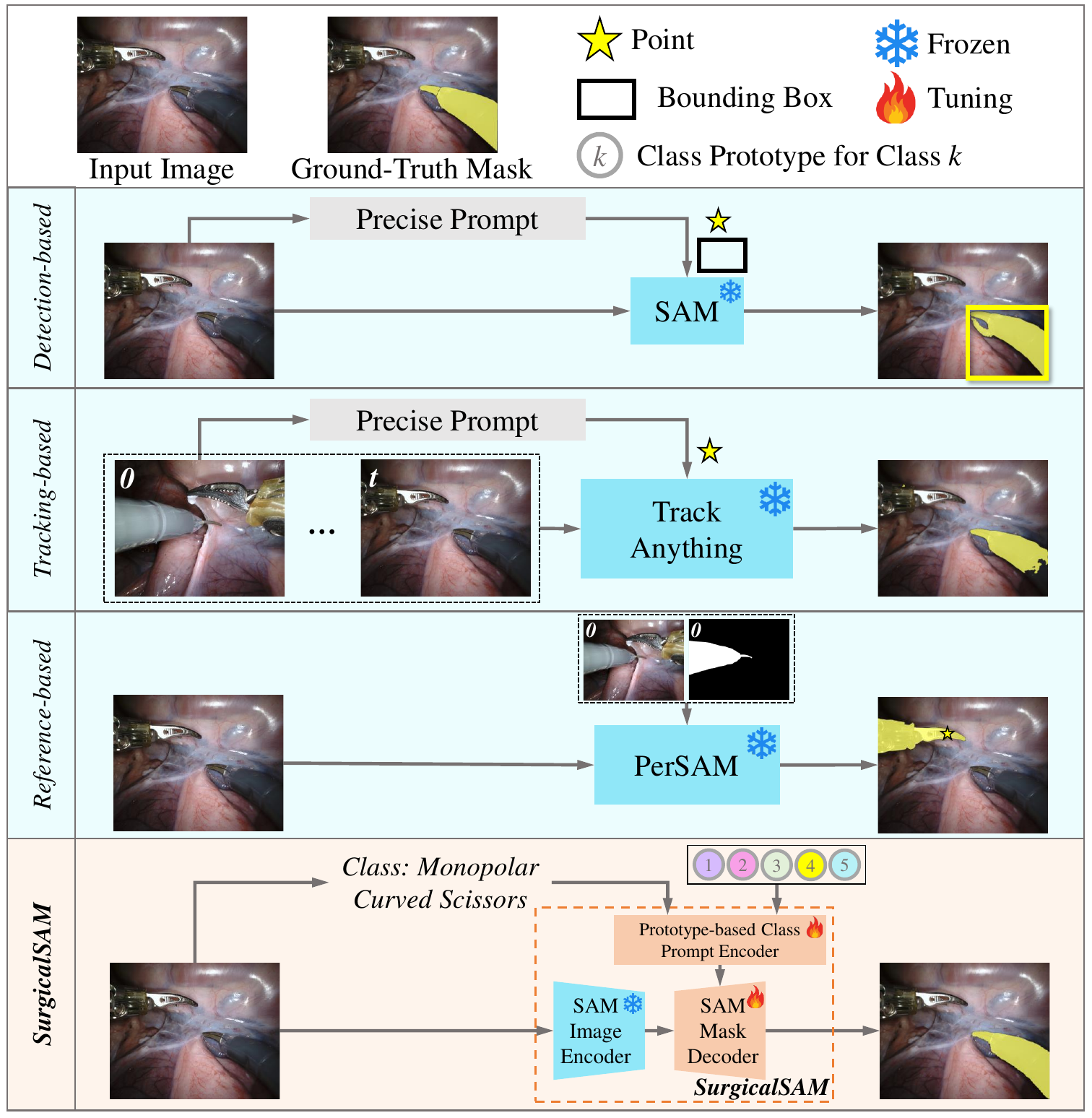}
\caption{Comparison of our SurgicalSAM against existing detection-based, tracking-based, and reference-based zero-shot SAM frameworks for surgical instrument segmentation. } 
\label{fig:motiv_compare}
\end{figure}

The Segment Anything Model (SAM) \cite{sam} has recently gained significant attention as a pioneering foundation model for promptable segmentation. 
Utilising SAM for downstream medical tasks holds great promise for enhancing training efficiency and leveraging strong pre-trained knowledge. 
Current research predominantly employs SAM in a zero-shot manner for medical image segmentation. However, the lack of sufficient medical data in SAM pre-training and the substantial domain gap between natural objects and medical targets hinders the direct generalisation of SAM towards medical tasks. Many studies have reported subpar performance of SAM in zero-shot medical image segmentation \cite{sam_wsi, sam_12datasets,sam_md,sam_mia, sam_yangxin, sam_ppt_modes,sam_surgical, sam_surgical_miccaiw}. 

\begin{figure}[!t]
    \centering
    \subfloat[SAM Prediction Mask mAP vs. Bounding Box Prompt Jitter]{
      \includegraphics[width=0.45\textwidth]{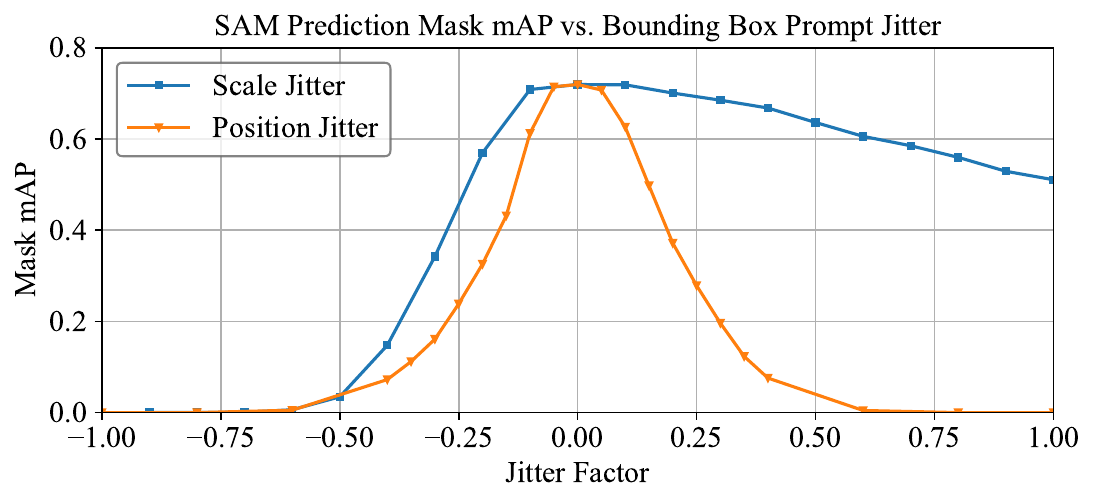}
    }   
    
    \subfloat[Scale Jitter -0.2]{
      \includegraphics[width=0.15\textwidth]{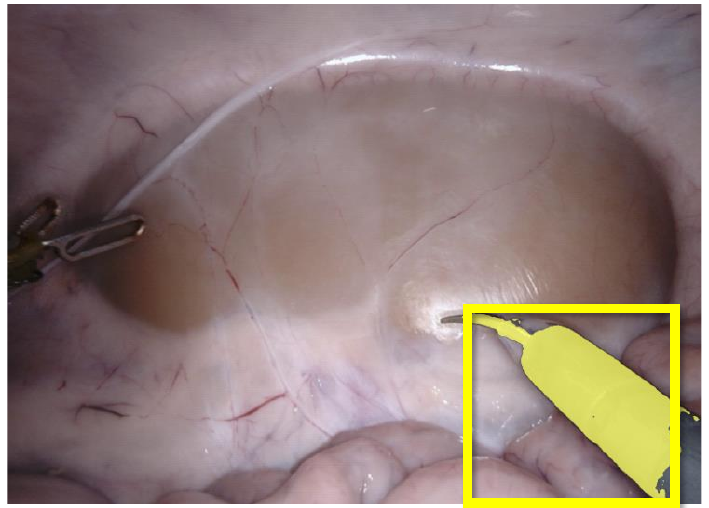}
    }
    \subfloat[GT Bounding Box]{
      \includegraphics[width=0.15\textwidth]{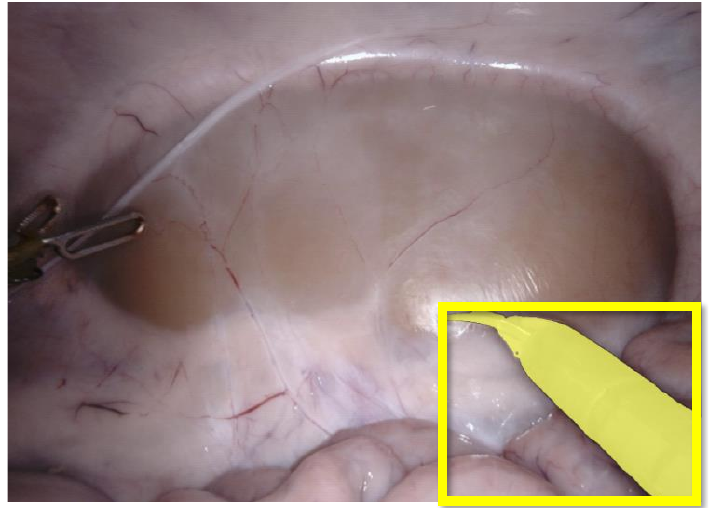}
    }
    \subfloat[Scale Jitter 0.4]{
      \includegraphics[width=0.15\textwidth]{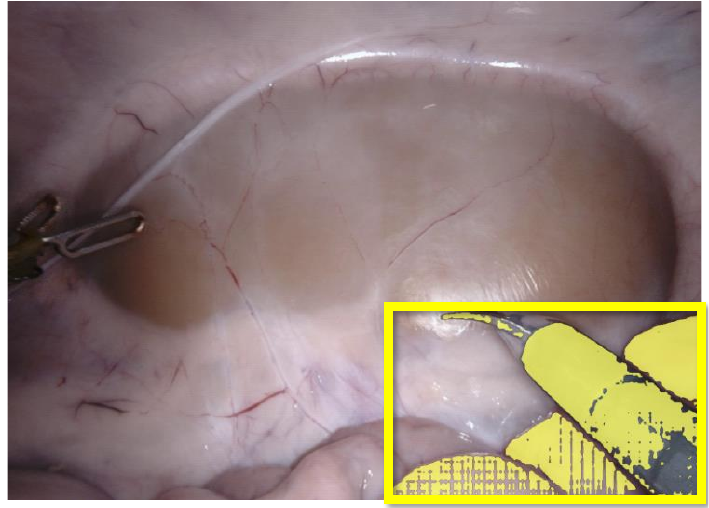}
    }

    \subfloat[Position Jitter -0.2 ]{
      \includegraphics[width=0.15\textwidth]{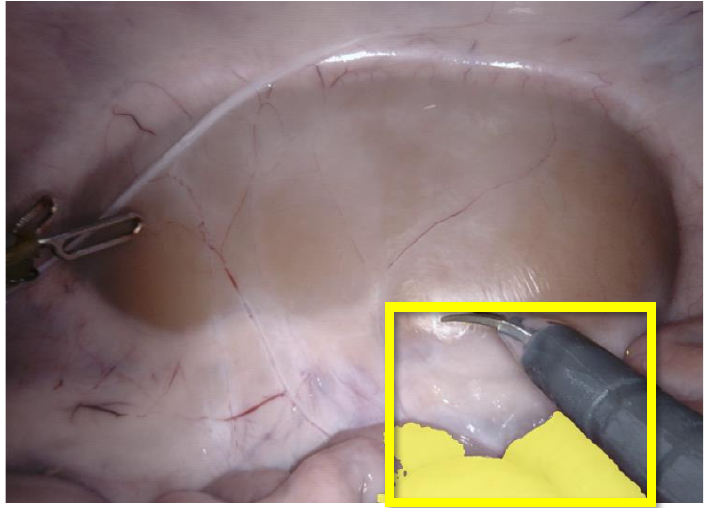}
      
    }
    \subfloat[GT Mask]{
      \includegraphics[width=0.15\textwidth]{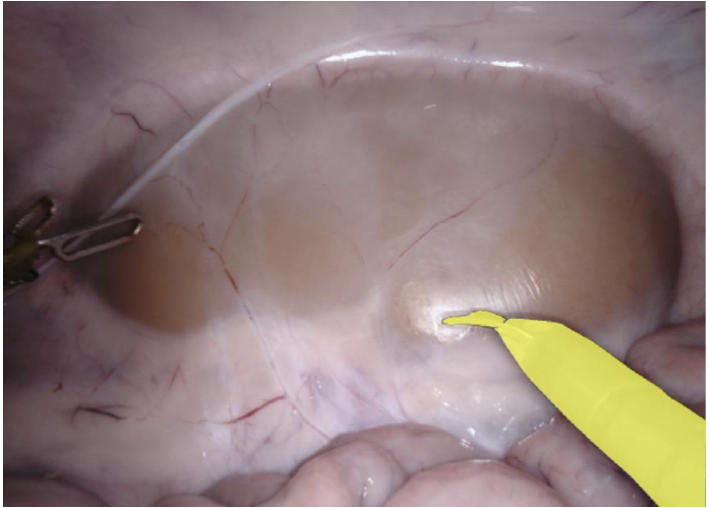}
    }
    \subfloat[Position Jitter 0.4]{
      \includegraphics[width=0.15\textwidth]{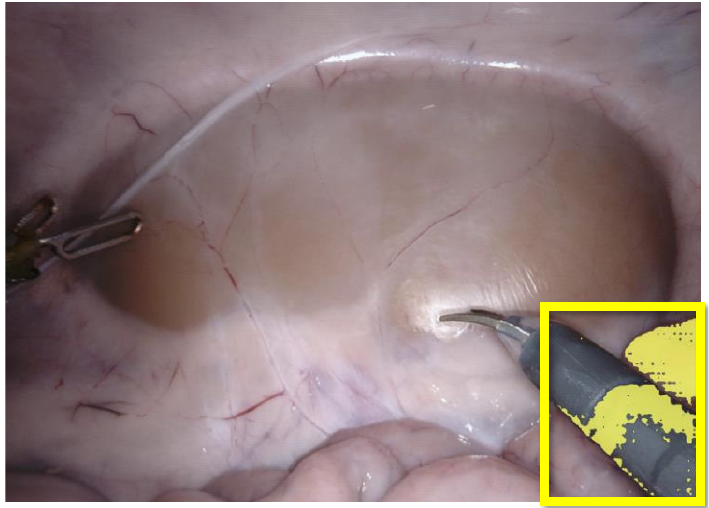}
    }
  
    \caption{Prompt robustness study of SAM against bounding box jitter in terms of scale and position for surgical instrument segmentation. A jitter factor of 0 represents the ground-truth bounding box with no jitter; a higher absolute value of the jitter factor indicates larger prompt noises.}
   \label{fig:robustness}
 \end{figure}

Specifically, surgical instruments differ significantly from natural objects in terms of specialised appearance, complex anatomical background, and high inter-category similarity. We evaluate three essential zero-shot SAM strategies on SIS: (1) MaskTrackRCNN \cite{masktrackrcnn} or Mask2Former \cite{m2f} as a bounding box detector followed by SAM, (2) Track Anything \cite{trackanything}, and (3) PerSAM \cite{persam}, representing detection-based, tracking-based, and reference-based frameworks, respectively. As shown in Fig. \ref{fig:motiv_compare}, these methods demonstrate inferior results, where detection-based and tracking-based methods depict incorrect contours and the reference-based method misidentifies the instrument class. This further highlights the challenge of bridging the natural-surgical domain gap and emphasises the necessity of SAM tuning.  

In addition, the performance of SAM relies on the precise locations of explicit prompts \cite{sam_ppt_modes, sam_md}. We confirm this through a prompt robustness study on SIS by introducing various scale and position jitters to the ground-truth bounding box as a prompt for SAM and recording the prediction mAP. As shown in Fig. \ref{fig:robustness}, our study demonstrates SAM's sensitivity to prompt jitters: even minor deviations in the provided bounding box prompts can significantly impair segmentation accuracy. 
As a result, existing zero-shot SAM frameworks often involve complex multi-stage pipelines, requiring either precise manual guidance or a well-performing specialist detector to provide accurate points or bounding boxes for accurate prompting. This complexity further restricts the direct application of SAM in the surgical domain.

To address the above challenges, we propose SurgicalSAM, an end-to-end approach that effectively mitigates the surgical-natural domain gap through efficient tuning of SAM.
A comparison of SurgicalSAM against existing pipelines is shown in Fig. \ref{fig:motiv_compare}. We propose a lightweight prototype-based class prompt encoder, which takes an instrument class as a prompt and learns the class prototypes by interacting with the image embedding to directly generate prompt embeddings for the mask decoder. By tuning the prototype-based class prompt encoder and the mask decoder, surgical knowledge is integrated with SAM's pre-trained knowledge, effectively mitigating the domain gap. Moreover, our strategy of directly generating latent prompt embeddings from class prompts and eliminating the use of explicit points and bounding boxes further addresses the poor robustness associated with explicit prompts as well as maintains an end-to-end pipeline.

In SurgicalSAM, the class prototypes play a vital role in effectively prompting the instrument of interest from an image. However, different surgical instrument categories often exhibit high similarity and low inter-class differences, thus posing a big challenge. To address this, we further propose contrastive prototype learning, utilising contrastive loss to acquire discriminative learned class prototypes. This method enhances the distinction between fine-grained instrument categories, resulting in more accurate class prompting and improved segmentation outcomes.

In summary, the contributions of this paper are threefold:
\begin{itemize}
    \item We introduce SurgicalSAM to integrate surgical instrument knowledge with the pre-trained knowledge in SAM through efficient tuning for class promptable surgical instrument segmentation. It outperforms both specialist models and complex multi-stage solutions.
    \item We propose a prototype-based class prompt encoder that eliminates the use of explicit prompts and facilitates direct learning of latent prompt embeddings from class prompts for an end-to-end pipeline. We also propose contrastive prototype learning to enhance the discrimination of the prototypes of fine-grained instrument categories for more accurate class prompting.
   \item We conduct extensive experiments on the challenging EndoVis2018 and EndoVis2017 datasets, achieving state-of-the-art (SOTA) performance while significantly improving training efficiency.
\end{itemize}

\section{Related Work}
\subsection{Surgical Instrument Segmentation}
Current research addresses SIS by training customised specialist models. Early research employs a pixel classification paradigm to predict pixel-wise class probabilities in a frame. Notably, TernausNet pioneers this direction using a U-Net-based encoder-decoder network \cite{ternausnet}. This has been later extended with feature pyramid attention \cite{paanet} and flow-based temporal priors \cite{mftapnet, dual_mf}.  Nevertheless, these approaches encounter spatial class inconsistency, where one instrument may be assigned multiple instrument types. 

An alternative paradigm is mask classification, which aims to predict a set of masks and associate each mask with a class label, inherently reducing spatial class inconsistency. 
ISINet introduces mask classification to instrument segmentation with Mask-RCNN \cite{isinet, mask_rcnn}. 
Later, \citet{baby} improve its classification performance by designing a specialised classification module. In addition, TraSeTR integrates tracking cues with a track-to-segment transformer \cite{trasetr} and MATIS incorporates temporal consistency with Mask2Former \cite{matis, m2f}.
Although various methods have been proposed for surgical instrument segmentation, they primarily rely on designing specialist models and training the complete set of model parameters, which is inefficient. Particularly with the small datasets in the surgical domain, these models may exhibit subpar generalisation performance.

\subsection{Segment Anything Model}
SAM is recognised as a pioneering foundation model for image segmentation. The large-scale pre-training equips it with excellent zero-shot generalisation capabilities, driving various downstream applications \cite{samrs, refsam, SAM_rs}. 
However, SAM has been shown to struggle with zero-shot generalisation to medical scenarios \cite{sam_wsi, sam_12datasets, sam_mia, sam_yangxin, sam_ppt_modes} due to the substantial domain gap between natural objects and medical subjects. Moreover, SAM relies on explicit points and bounding boxes at precise locations for accurate segmentation \cite{sam_ppt_modes, sam_md}. 
As a result, extensive manual guidance or a specialist detector is often required, leading to a complex multi-stage pipeline \cite{sam_surgical}.  

To bridge the natural-medical domain gap, some studies seek to adapt SAM through domain-specific fine-tuning. However, they either require accurate point or bounding box prompts \cite{medsam, msa} or employ universal prompt embeddings for all classes which lack discrimination for fine-grained surgical instrument categories \cite{sam_ada_cust, sam_ada_udpf, sam_surgical_miccaiw}. In contrast, our approach introduces a novel efficient-tuning approach for SAM with a prototype-based prompt encoder, which generates prompt embeddings from contrastively-learned class prototypes. This enhances the discrimination of fine-grained classes while simplifying the pipeline by eliminating the need for explicit prompts.

\section{Methodology}

\subsection{Overview}
In this work, we address the task of surgical instrument segmentation in a class promptable manner through efficient tuning of SAM. Specifically, given a surgical image $I \in \mathbb{R}^{H \times W \times 3}$ with spatial resolution $H \times W$ and the class of an instrument in the image $c$ as prompt, our goal is to predict the class $c$ mask of the image, denoted as $M^{(c)}$:
\begin{equation}
M^{(c)} = SurgicalSAM (I, c).
\end{equation}

\begin{figure*}[!t]
    \centering
    \subfloat[Overview of SurgicalSAM]{
      \includegraphics[width=0.65\textwidth]{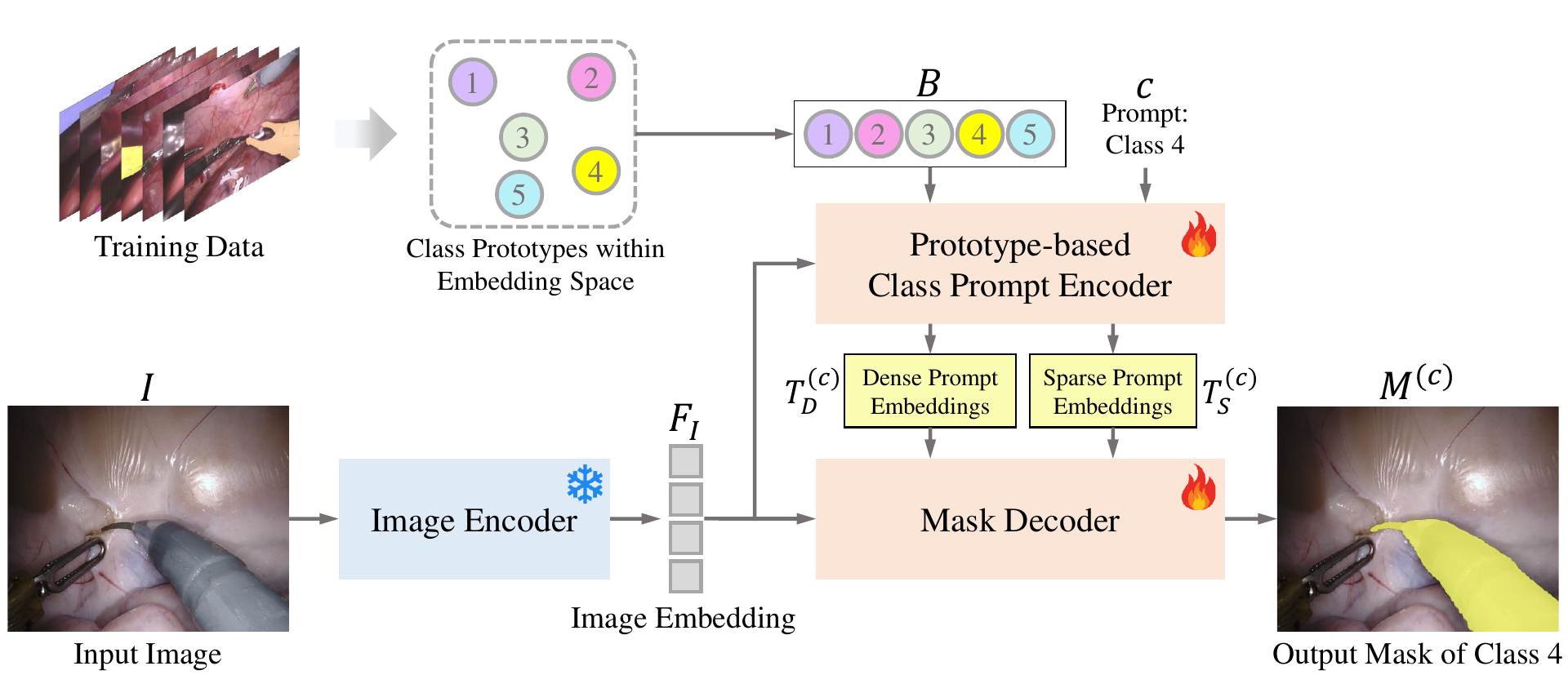}
    }
    \hfill
    \subfloat[Prototype-based Class Prompt Encoder]{
      \includegraphics[width=0.3 \textwidth]{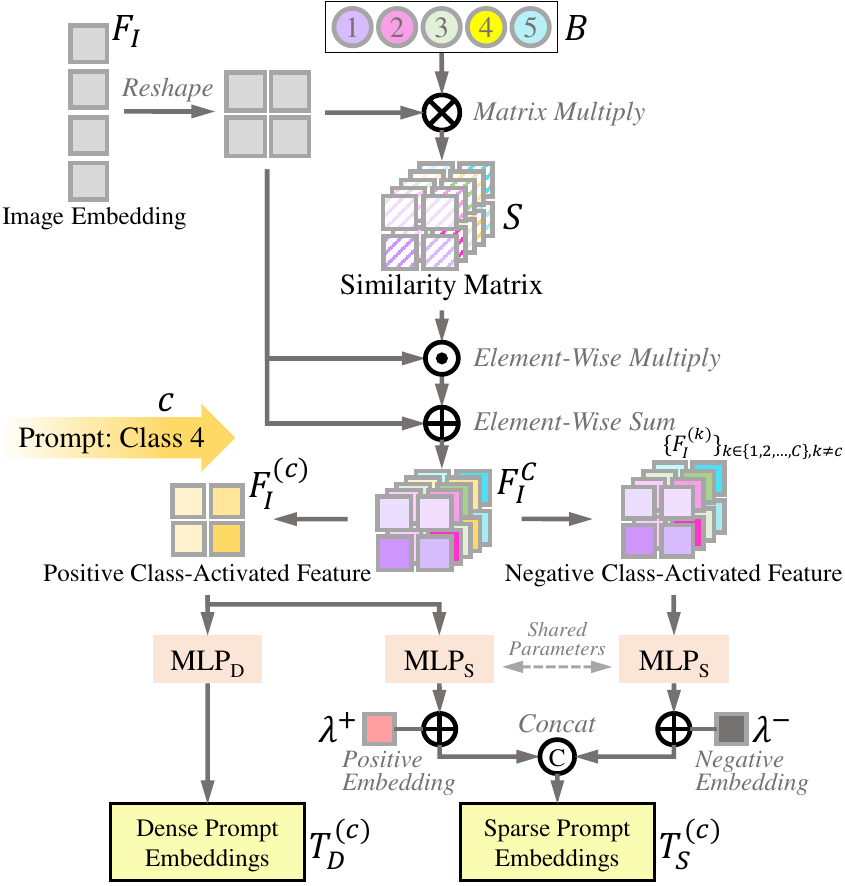}
    }
    \caption{SurgicalSAM for class promptable surgical instrument segmentation through efficient tuning of SAM. }
   \label{fig:method_overview}
 \end{figure*}

SurgicalSAM is composed of three core components as shown in Fig. \ref{fig:method_overview}(a): an image encoder, a prototype-based class prompt encoder, and a mask decoder. Similar to SAM, the image encoder $E_I$ first extracts the embedding of the input image as $F_I \in \mathbb{R}^{h \times w \times d}$, with $h \times w$ denoting the shape of the image embedding and $d$ representing the number of embedding channels. Then, our prototype-based class prompt encoder $E_{CP}$ utilises the class prototypes $B$ to activate the image embedding and leverages the obtained activated feature conditioned on the prompt class $c$ to generate prompt embeddings, including dense prompt embeddings $T_{D}^{(c)}$ and sparse prompt embeddings $T_{S}^{(c)}$. Finally, the image embedding and prompt embeddings are used to predict the mask $M^{(c)}$ by the mask decoder $D_M$. The above process can be expressed as:
\begin{equation}
F_I = E_I (I),
\end{equation}
\begin{equation}
T_{D}^{(c)}, T_{S}^{(c)} = E_{CP} (F_I, B, c),
\end{equation}
\begin{equation}
M^{(c)} = D_M (F_I, [T_{D}^{(c)}, T_{S}^{(c)}, T_O]),
\end{equation}
where $T_O$ denotes the learnable output tokens in SAM.

\subsection{Prototype-based Class Prompt Encoder}

The prototype-based class prompt encoder exploits the similarity between the image and class prototypes to create prompt embeddings. 
Specifically, as shown in Fig. \ref{fig:method_overview}(b), the spatial-wise similarity between the image embedding and the class prototype is computed to activate class-specific regions within the image, resulting in a class-activated feature to generate prompt embeddings for the mask decoder. Furthermore, inspired by the utilisation of both foreground and background point prompts in SAM, we propose to not only employ the prototype of the prompted class but integrate all class prototypes to incorporate both positive and negative cues. 
Such a strategy provides more robust priors for the model to effectively distinguish between instrument classes with high similarity. 

Specifically, the prototype-based class prompt encoder $E_{CP}$ is built upon a prototype bank $B = concat(\{B^{(k)}\}_{k \in \{1,2,...,C\} }) \in \mathbb{R}^{C \times d}$ consisting of a representative prototype for each class, where $C$ is the total number of classes. Given an image $I$  with image embedding $F_I$, we construct a similarity matrix $S = concat(\{S^{(k)}\}_{k\in\{1,2,...,C\}})\in \mathbb{R}^{C \times h \times w}$ to represent the spatial-wise similarity of the image with the prototypes of all classes. It is generated by computing the dot product between the image embedding at every spatial location and each class prototype:
\begin{equation}
S^{(k)} = F_I \times B^{(k)}, \text{for } k \in \{1,2, ..., C\}.
\end{equation}
The similarity matrix is then employed as spatial attention to activate the class-specific regions, resulting in class-activated feature for all classes $F_I^C = concat(\{ F_I^{(k)}\}_{k \in \{1,2,...,C\}}) \in  \mathbb{R}^{C \times h \times w \times d}$: 
\begin{equation}
    F^{(k)}_I = F_I \circ S^{(k)} + F_I , \text{for } k \in \{1,2, ..., C\},
\end{equation}
where $\circ$ and $+$ represents element-wise multiplication and addition, respectively, and $F_I^{(k)} \in \mathbb{R}^{h \times w \times d}$ represents the class-activated feature for class $k$.

Finally, the class-activated feature is used to formulate dense and sparse prompt embeddings. 
In SAM, dense prompt embeddings are derived from foreground masks, providing \textit{positive} cues for segmenting the object. Imitating this, we leverage the class-activated feature of the \textit{positive} class, \ie, the prompted class $c$, for encoding dense prompt embeddings $T_D^{(c)}\in \mathbb{R}^{h \times w \times d}$.  
This is achieved through a two-layer Multilayer Perceptron (MLP): 
\begin{equation}
T_D^{(c)} = g_D(ReLU(f_D(F_I^{(c)}))),
\end{equation}
where $f_D$ and $g_D$ are two linear projection functions with intermediate dimension $r_D$. On the other hand, the sparse prompt embeddings in SAM are encoded from both \textit{positive} information (foreground points and bounding boxes) and \textit{negative} information (background points). Inspired by this, we generate sparse prompt embeddings using the class-activated feature of all classes that include both \textit{positive}, prompted class and \textit{negative}, non-prompted classes. The positive and negative classes are then distinguished through a pair of positive and negative embeddings. Specifically, $F_I^C$ is first fed into a two-layer MLP to obtain positivity-agnostic sparse prompt embeddings $\hat{T}_S^C = concat(\{ \hat{T}_S^{(k)}\}_{k \in \{1,2,...,C\}}) \in \mathbb{R}^{C \times n \times d}$:
\begin{equation}
\hat{T}_S^C = g_S(ReLU(f_S(F_I^C))),
\end{equation}
where $f_S$ and $g_S$ are two linear projection functions with intermediate dimension $r_S$, $n$ indicates the number of sparse tokens per class, and $\hat{T}_S^{(k)} \in \mathbb{R}^{n\times d}$ represents the positivity-agnostic sparse prompt embedding activated by class $k$. Then, a pair of positive and negative embeddings, $\lambda^{+} \in \mathbb{R} ^d$ and $\lambda^{-}\in \mathbb{R} ^d$, are respectively added to the embeddings corresponding to positive class (class $c$) and negative classes (classes other than $c$), resulting in the final sparse prompt embeddings ${T}_S^{(c)} \in \mathbb{R}^{C \times n \times d}$ that are positivity-aware:
\begin{align}
    {T}_S^{(c)} &= concat(\{ \hat{T}_S^{(k)} + \mathds{1}(k=c) \lambda^{+} + (1-\mathds{1}(k=c)) \lambda^{-} \}), \nonumber \\
    & \ \ \quad \text{for }{k \in \{1,2,...,C\}}.
\end{align}
$T_S^{(c)}$ is then reshaped to $Cn \times d$ and is fed with $T_D^{(c)}$ into the mask decoder for mask prediction.

\subsection{Contrastive Prototype Learning}
\begin{figure}[!t]
\centering
\includegraphics[width=0.45\textwidth]{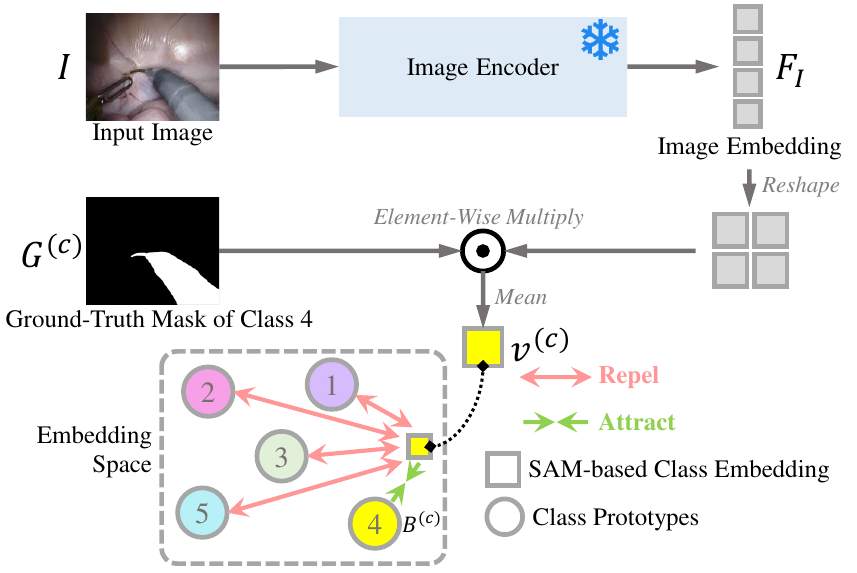}
\caption{Contrastive Prototype Learning.} 
\label{fig:cl}
\end{figure}

Our method relies on discriminative class prototypes for precise instrument category identification and accurate class region activation. However, obtaining accurate class prototypes in surgical scenarios with highly similar instrument appearances is challenging. To enhance prototype discriminativeness for more accurate class prompting, we propose contrastive prototype learning to acquire the optimised class prototypes during tuning of the framework, as illustrated in Fig. \ref{fig:cl}. Specifically, we propose prototype contrastive loss motivated by infoNCE loss \cite{infonce, mutual}, where the class prototypes are considered as anchors and the SAM-based class embeddings in training images are regarded as samples. Given image embedding $F_I$, the ground-truth binary mask of class $c$, $G^{(c)}$, is processed to resolution $h \times w$ and used to extract the SAM-based class embedding $v^{(c)} \in \mathbb{R} ^{d}$ for class $c$ by averaging the foreground features:
\begin{equation}
    v^{(c)} = \frac{\sum_i^{hw} (F_I \circ G^{(c)})}{\sum_i^{hw} G^{(c)}}.
\end{equation}

To this end, the prototype contrastive loss is expressed as:  
\begin{equation}
    \mathcal{L}_{PCL} =  - \frac{1}{C}\sum^{C}_{k=1} log\frac{exp(B^{(k)}\cdot v^{(k)}/\tau)}{\sum_{q=1}^C exp(B^{(k)}\cdot v^{(q)}/\tau)}, \label{eq:MI_lower}
\end{equation}
where $\tau$ refers to the temperature parameter for modulating the similarities and $B^{(k)}$ is the prototype of class $k$. It can be seen that $\mathcal{L}_{PCL}$ strengthens the similarity between the prototype of class $k$ (anchor) and the SAM-based class embeddings of $k$ (positive samples), simultaneously suppressing the similarity between the prototype of class $k$ (anchor) with the SAM-based class embeddings of the classes other than $k$ (negative samples). This results in more discriminative prototype representations and enhanced surgical domain knowledge infusion through SAM tuning.

\subsection{Efficient Tuning}
SurgicalSAM is of high training efficiency. During tuning, the large image encoder is frozen and only the parameters of the lightweight prototype-based prompt encoder and mask decoder are updated. The tuning is end-to-end, supervised by a loss function consisting of two terms: dice loss for segmentation \cite{vnet} and prototype contrastive loss for prototype learning: 
\begin{equation}
    \mathcal{L} = \mathcal{L}_{DICE} + \mathcal{L}_{PCL},
\end{equation}
\begin{equation}
    \mathcal{L}_{DICE} = \frac{2 \sum_i^{HW} m_i g_i}{\sum_i^{HW} m_i^{2} + \sum_i^{HW} g_i^{2}},
\end{equation}
where $m_i$ and $g_i$ are the predicted logit value and the ground-truth binary value at pixel $i$ of the image, respectively.

\section{Experiments and Discussion}
\subsection{Datasets and Evaluation}
We validate our method using the EndoVis2018 \cite{endovis2018} and EndoVis2017 \cite{endovis2017} datasets.
For a fair comparison with existing methods, we adhere to the standard experiment and evaluation protocols defined by \citet{ternausnet} and \citet{isinet}. EndoVis2017 consists of eight videos, each with 255 frames, for which we perform 4-fold cross-validation following the fold division provided by \citet{ternausnet}. 
EndoVis2018 offers 11 training videos and four validation videos with each consisting of 149 frames. Both datasets provide seven instrument categories.

For evaluation, we follow prior research and adopt three segmentation metrics: Challenge IoU \cite{endovis2017}, IoU, and mean class IoU (mc IoU) \cite{isinet, baby, matis}. 
The efficiency of our method is evaluated in terms of training speed, training GPU usage, and inference speed.

\subsection{Implementation Details}
The data from EndoVis2017 and EndoVis2018 are pre-processed following \citeauthor{ternausnet} \shortcite{ternausnet}. 
For the prototype-based prompt encoder, the intermediate dimensions $r_D$ and $r_S$ are both set to 128 and the number of tokens per class $n$ is set to 2 and 4 for EndoVis2018 and EndoVis2017, respectively. For prototype contrastive loss, a temperature $\tau$ of 0.07 is used. In terms of training, we initialise the image encoder, the mask decoder, and the positive and negative embeddings ($\lambda^+$ and $\lambda^-$) of SurgicalSAM with SAM's pre-trained weight of the ViT-H version \cite{vit}.  
The image encoder and the positive and negative embeddings of our model remain frozen while the weights of the prompt encoder and mask decoder are updated. 
We employ an Adam optimiser with a learning rate of 0.001 and 0.0001 for EndoVis2018 and EndoVis2017, respectively.
To reduce computational load, we adopt pre-computed image embeddings in training, employing a batch size of 32. Our model is implemented using PyTorch and trained and evaluated on an Nvidia Tesla V100 16GB GPU.

\subsection{Main Results}

\begin{table*}[h]
\centering
\renewcommand{\arraystretch}{1}
\setlength{\tabcolsep}{5pt}
\begin{adjustbox}{width=\textwidth}
\begin{tabular}{l l c c c c c c c c c c c}
\hline
   &  &  &  &  & \multicolumn{7}{c}{Instrument Categories} &  \\ \cline{6-12} 
 \multirow{-2}{*}{Method Category} & \multirow{-2}{*}{Method} & \multirow{-2}{*}{Challenge IoU} & \multirow{-2}{*}{IoU} & \multirow{-2}{*}{mc IoU} & BF & PF & LND & SI & CA & MCS & UP & \multirow{-2}{*}{\#Params}\\ \hline
\multirow{7}{*}{Specialist Model} & TernausNet & 46.22 & 39.87 & 14.19 & 44.20 & 4.67 & 0.00 & 0.00 & 0.00 & 50.44 & 0.00 & 32.20M \\
  & MF-TAPNet  & 67.87 & 39.14 & 24.68 & 69.23 & 6.10 & 11.68 & 14.00 & 0.91 & 70.24 & 0.57 & 37.73M \\
  & Dual-MF & 70.40 & - & 35.09 & 74.10 & 6.80 & 46.00 & 30.10 & 7.60 & 80.90 & 0.10 & 203.80M \\
   & ISINet &  73.03 & 70.94 & 40.21 & 73.83 & 48.61 & 30.98 & 37.68 & 0.00 & 88.16 & 2.16 & 162.52M\\
  & TraSeTr &  76.20 & -  & 47.71  & 76.30  & 53.30  & 46.50  & 40.60 & 13.90 & 86.20  & 17.15 & -  \\
    & S3Net & 75.81  &  74.02 &  42.58  & 77.22  &  50.87 &  19.83 &  50.59 &  0.00 &  92.12  & 7.44 & 68.41M  \\      
  & MATIS Frame & 82.37 & 77.01 & 48.65 & 83.35 & 38.82 & 40.19 & 64.49 & 4.32 & 93.18 & 16.17 & 68.72M\\\hline
\multirow{9}{*}{SAM-based Model}  & MaskTrack-RCNN + SAM & 78.49 & 78.49 & 56.07 & 79.83 & 74.86 & 43.12 & \textbf{62.88} & 16.74  &  91.62 & 23.45 & 57.67M  \\
 & Mask2Former + SAM & 78.72 & 78.72 & 52.50 & \textbf{85.95} & \textbf{82.31} & 44.08  & 0.00 & \textbf{49.80} & \textbf{92.17} & 13.18 & 68.72M \\
 & TrackAnything (1 Point)  &  40.36 &	38.38 & 20.62 & 30.20 & 12.87 & 24.46  & 9.17 & 0.19& 55.03 & 12.41 & -\\
 & TrackAnything (5 Points)  &  65.72 &  60.88 & 38.60 & 72.90 & 31.07 & \textbf{64.73}  & 10.24 & 12.28 & 61.05 & 17.93 & -\\
 & PerSAM &  49.21 & 49.21 & 34.55 & 51.26 & 34.40 & 46.75 & 16.45 & 15.07 & 52.28 & \textbf{25.62} & - \\
 & PerSAM (Fine-Tune) &  52.21 & 52.21 & 37.24 & 57.19 & 36.13 & 53.86 & 14.34 & 25.94 & 54.66 & 18.57 & 2 \\
 & \textbf{SurgicalSAM} & \textbf{80.33} & \textbf{80.33} & \textbf{58.87} & 83.66 & 65.63 & 58.75  & 54.48 & 39.78 & 88.56 & 21.23 & 4.65M\\ \cline{2-13} 
 & GT Centroid + SAM & 60.26 & 60.26 & 63.34 & 44.35 & 65.92 & 30.99 & 87.14 & 69.69 & 80.04 & 65.26  & -\\
 & GT Bbox + SAM & 88.04 & 88.04 & 84.23 & 87.10 & 86.81 & 72.23 & 91.21 & 75.91 & 93.08 & 83.24 & -\\
\hline
\end{tabular}
\end{adjustbox}
\caption{Comparative Results on the EndoVis2018 Dataset. \#Params represents number of tunable parameters.}
\label{tab:results_endovis2018}
\end{table*}

\begin{table*}[h]
\centering
\renewcommand{\arraystretch}{1}
\setlength{\tabcolsep}{7.5pt}
\begin{adjustbox}{width=\textwidth}
\begin{tabular}{l l c c c c c c c c c c}
\hline
   &  &  &  &  & \multicolumn{7}{c}{Instrument Categories} \\ \cline{6-12} 
 \multirow{-2}{*}{Method Category} & \multirow{-2}{*}{Method} & \multirow{-2}{*}{Challenge IoU} & \multirow{-2}{*}{IoU} & \multirow{-2}{*}{mc IoU} & BF & PF & LND & VS & GR & MCS & UP \\ \hline
\multirow{7}{*}{Specialist Model} & TernausNet & 35.27 & 12.67 &  10.17 & 13.45 & 12.39 & 20.51 & 5.97 & 1.08 & 1.00 & 16.76 \\
  & MF-TAPNet & 37.25 & 13.49 & 10.77 & 16.39 &  14.11 & 19.01&  8.11 &  0.31   & 4.09  & 13.40 \\
   & Dual-MF & 45.80 & - & 26.40 & 34.40 & 21.50 & 64.30 & 24.10 & 0.80 & 17.90 & 21.80 \\
  & ISINet & 55.62 & 52.20 & 28.96 & 38.70 & 38.50 & 50.09 & 27.43 & 2.10 & 28.72 & 12.56 \\ 
  & TraSeTr & 60.40 & - & 32.56 & 45.20  &   56.70  &  55.80 &  38.90  &  11.40 &  31.30 &  18.20  \\
    & S3Net & 72.54 &  71.99 & 46.55 & 75.08
 & 54.32  & 61.84 & 35.50 &  27.47  & 43.23  & 28.38 \\      
  & MATIS Frame & 68.79 & 62.74 & 37.30 & 66.18 & 50.99 & 52.23 & 32.84 & 15.71 & 19.27 & 23.90 \\\hline
\multirow{8}{*}{SAM-based Model}  &  Mask2Former + SAM & 66.21 & 66.21 & 55.26 & 66.84 & \textbf{55.36} & \textbf{83.29} & 73.52 & 26.24 & 36.26 & 45.34 \\
 & TrackAnything (1 Point) & 54.90 & 52.46 & 55.35 & 47.59 & 28.71 & 43.27 & 82.75 & \textbf{63.10} & 66.46 & 55.54  \\
& TrackAnything (5 Points) & 67.41 & 64.50 & 62.97 & 55.42 & 44.46 & 62.43 & \textbf{83.68} & 62.59 & 67.03 & \textbf{65.17} \\
 & PerSAM & 42.47 & 42.47 & 41.80 & 53.99 & 25.89 & 50.17 & 52.87 & 24.24 & 47.33 & 38.16 \\
& PerSAM (Fine-Tune) & 41.90 & 41.90 & 39.78 & 46.21 & 28.22 & 53.12 & 57.98 & 12.76 & 41.19 & 38.99\\
 & \textbf{SurgicalSAM} & \textbf{69.94} & \textbf{69.94} & \textbf{67.03} & \textbf{68.30} & 51.77 & 75.52 & 68.24 & 57.63 & \textbf{86.95} & 60.80
 \\ \cline{2-12}
 & GT Centroid + SAM & 44.42 & 44.42 & 54.41 & 63.42 & 36.03 & 22.57 & 54.21 & 75.18 & 70.17 & 59.25 \\
 & GT Bbox + SAM & 76.31 & 76.31 & 81.18 & 89.36 & 73.44 & 67.67 & 90.04 & 87.79 & 94.03 & 65.91 \\
\hline
\end{tabular}
\end{adjustbox}
\caption{Comparative Results on the EndoVis2017 Dataset.}
\label{tab:results_endovis2017}
\end{table*}

\begin{figure}[!t]
\centering
\includegraphics[width=0.45\textwidth]{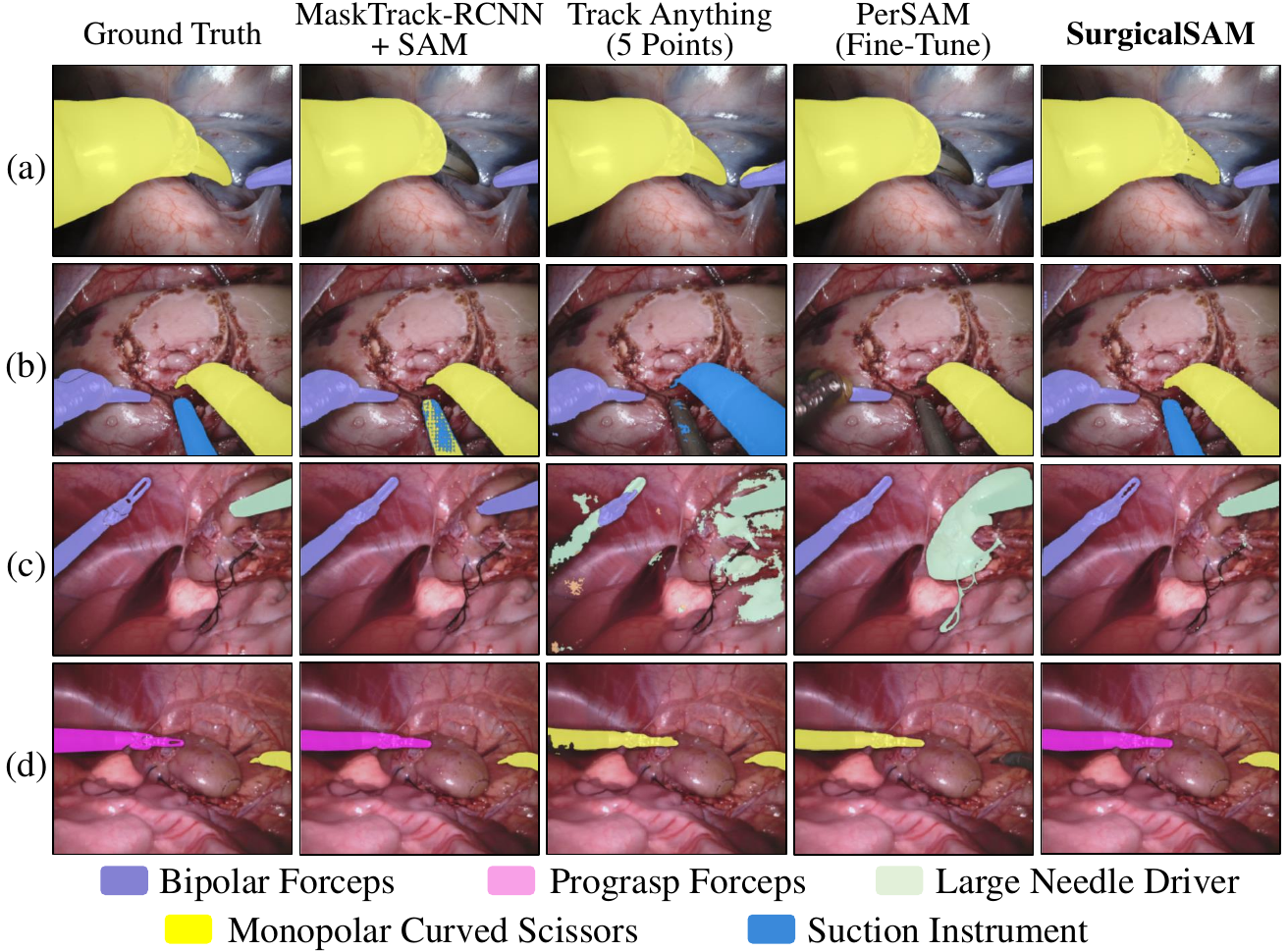}
\caption{Visualisation of Predicted Masks.} 
\label{fig:vis_results}
\end{figure}

The results of SurgicalSAM on EndoVis2018 and EndoVis2017 in comparison with existing methods are presented in Table \ref{tab:results_endovis2018} and Table \ref{tab:results_endovis2017}, respectively. A visual comparison of the predictions is shown in Fig. \ref{fig:vis_results}. 
The evaluated instrument categories include Bipolar Forceps (BF), Prograsp Forceps (PF), Large Needle Driver (LND), Suction Instrument (SI), Vessel Sealer (VS), Clip Applier (CA), Grasping Retractor (GR), Monopolar Curved Scissors (MCS), and Ultrasound Probe (UP). 
In our comparison, we categorise existing strategies into specialist models and SAM-based models. Remarkably, SurgicalSAM surpasses existing SAM-based models, matching or even exceeding the performance of SOTA specialist models, while using only a few tunable parameters.

In terms of SAM-based models, the three zero-shot SAM baselines: MaskTrack-RCNN or Mask2Former with SAM  \cite{masktrackrcnn,m2f} (detection-based), Track Anything \cite{trackanything} (tracking-based), and PerSAM \cite{persam} (reference-based), all exhibit inferior performance. 
In particular, PerSAM is notably unsuitable for the task due to its reliance on a single instance for visual reference and a simple two-point prompting mechanism. Given the substantial intra-class variance and low inter-class variance among surgical instruments, a single instance lacks the necessary information for accurately referencing an instrument, resulting in missing instances in prediction, as shown in Fig. \ref{fig:vis_results}(b) and (d). Additionally, the use of just one foreground point and one background point fails to effectively prompt SAM for zero-shot instrument segmentation due to SAM's lack of surgical domain knowledge, leading to an incorrect interpretation of the instrument contours (Fig. \ref{fig:vis_results}(a), (b), and (c)).
While Track Anything exhibits improved performance compared to PerSAM, its efficacy heavily relies on the quality of prompts, as shown by the large gap between the results obtained from prompting with one point versus five points. Furthermore, the significant motion of instruments often causes Track Anything to lose track or confuse between instruments with similar appearances (Fig. \ref{fig:vis_results}(b), (c), and (d)). 
Detection-based SAM shows the most promising performance among the three zero-shot SAM baselines. However, its effectiveness relies on a well-trained detector model which requires significant training effort. Also, without SAM tuning, the lack of domain knowledge can result in incomplete masks or misidentification of instrument categories (Fig. \ref{fig:vis_results}(a), (b), and (c)). 

Our SurgicalSAM outperforms {\it all} three zero-shot SAM baselines. Different from these solutions, it integrates surgical domain knowledge with SAM's pre-trained general knowledge, enhancing its expertise with surgical instruments and resulting in more accurate segmentation (Fig. \ref{fig:vis_results}). Meanwhile, the tuning of SurgicalSAM is highly efficient, requiring significantly fewer tunable parameters than the detection-based model (SurgicalSAM with 4.65M parameters vs. MaskTrack-RCNN + SAM with 57.67M parameters). Furthermore, SurgicalSAM utilises learned prototypes as references, which are more general and descriptive than the single instance reference in PerSAM, and eliminates the use of explicit prompts for a pipeline much simpler than the multi-stage detection-based pipeline.

We also establish two oracle scenarios by employing ground-truth centroids or ground-truth bounding boxes as prompts for SAM. As shown in Table \ref{tab:results_endovis2018} and Table \ref{tab:results_endovis2017}, SurgicalSAM demonstrates substantial superiority over the utilisation of ground-truth centroids, achieving an improvement of 20.07\% and 25.52\% in Challenge IoU for EndoVis2018 and EndoVis2017, respectively. These promising results show that SurgicalSAM already attains superior results compared to employing basic manual guidance. 

Last but not least, our method achieves SOTA performance competitive with the specialist models on both datasets, while requiring substantially fewer tunable parameters (SurgicalSAM with 4.65M parameters vs. MATIS Frame with 68.72M parameters). Particularly, significant improvements can be observed in mean class IoU, indicating that the general knowledge in foundation models serves as extra priors that help to diminish the class imbalance problem in small datasets.  
In summary, our method achieves promising performance with high efficiency. 

\begin{table}[]
\renewcommand{\arraystretch}{1}
\setlength{\tabcolsep}{8.5pt}
\centering
\begin{adjustbox}{width=0.46\textwidth}
\begin{tabular}{c c c c c }
\hline
  & Challenge IoU & mc IoU & Challenge IoU & mc IoU  \\  \hline
 $n$ \textbackslash $\mathcal{L}_{PCL}$ & \multicolumn{2}{c}{\ding{55}} &  \multicolumn{2}{c}{\ding{51}} \\ \hline
2  & 76.38 &  53.95 & \textbf{80.33} & \textbf{58.87}    \\ 
4  & 78.26 & 56.54 & 79.46  & 58.40   \\ 
6  & 77.28 & 53.71  &  79.67  &  56.97  \\ 
8  & 76.98 & 53.94 & 80.10 & 58.30 \\
\hline
\end{tabular}
\end{adjustbox}
\caption{Ablation Study on SurgicalSAM.}
\label{tab:ablat}
\end{table}

\subsection{Ablation Study}
We conduct an ablation study on EndoVis2018 for contrastive prototype learning and the number of tokens $n$. Specifically, we remove the contrastive prototype learning module and use fixed class prototypes computed by taking the average of the class embeddings across all training samples. The results, as depicted in Table \ref{tab:ablat}, show a significant difference. Without the contrastive learning process, the pre-computed fixed prototypes tend to be overly similar across different instrument categories due to their highly similar appearance. Contrastive prototype learning helps the model to learn more discriminative class prototypes and accurately identify the instrument classes. Moreover, the efficacy of contrastive prototype learning remains consistent across different numbers of tokens. Regarding the impact of different numbers of tokens on our complete model, as shown in Table \ref{tab:ablat}, no notable changes can be observed. In contrast to the original SAM which is sensitive to the number of points provided \cite{sam_ppt_modes}, the use of class prompt in our work demonstrates enhanced robustness.

\subsection{Cross-Dataset Generalisation}

We verify the cross-dataset generalisability of SurgicalSAM by training it on one dataset and evaluating it on another. The results are shown in Table \ref{tab:cross_data}, where only the instrument classes shared by both datasets are considered. Compared to the SOTA specialist model MATIS Frame, our method consistently performs better in both ways (EndoVis2018 to EndoVis2017 and EndoVis2017 to EndoVis2018). Notably, when trained on EndoVis2018 and evaluated on EndoVis2017, we achieve a large improvement of 11.43\% in the IoU averaged over all classes. This underscores the advantage of SurgicalSAM over dedicated specialist models in terms of its ability to effectively generalise to new data distributions, owing to its integration of both foundation general knowledge and surgical domain expertise.

\begin{table}[]
\renewcommand{\arraystretch}{1}
\centering
\begin{adjustbox}{width=0.46\textwidth}
\begin{tabular}{c c l c c c c c c}
\hline
 &  &  & \multicolumn{4}{c}{Instrument Categories (IoU)} &   \\  \cline{4-7}
  \multirow{-2}{*}{\textit{T}}  & \multirow{-2}{*}{\textit{V}}  & \multirow{-2}{*}{Method}  & BF & PF & LND & MCS & \multirow{-2}{*}{Mean IoU}  \\  \hline
\multirow{2}{*}{\textit{18}} & \multirow{2}{*}{\textit{17}} & MATIS Frame & 45.57 & 32.62 & 44.98 & 58.84 & 45.50\\
  & & \textbf{SurgicalSAM} & \textbf{70.95} & \textbf{35.21} & \textbf{45.46} & \textbf{76.08} & \textbf{56.93} \\ \hline 
\multirow{2}{*}{\textit{17}} & \multirow{2}{*}{\textit{18}} & MATIS Frame & \textbf{65.55} & 13.89 & 38.25 & \textbf{65.58} &  45.81\\
  & &  \textbf{SurgicalSAM} & 44.50 & \textbf{27.17} & \textbf{50.76} & 62.94 & \textbf{46.34} \\ \hline 
\end{tabular}
\end{adjustbox}
\caption{Cross-Dataset Generalisation. \textit{T}: training dataset; \textit{V}: validation dataset; \textit{18}: EndoVis2018; \textit{17}: EndoVis2017.}
\label{tab:cross_data}
\end{table}

\begin{table}[]
\renewcommand{\arraystretch}{1}
\centering
\begin{adjustbox}{width=0.46\textwidth}
\begin{tabular}{l c c c c c c}
\hline
  & \multicolumn{3}{c}{Training Speed (fps)}  & \multicolumn{3}{c}{Training GPU Usage (GB)} \\ \cline{2-7}  
 \multirow{-2}{*}{Method}  & bz = 2   & bz = 16  & bz = 32 & bz = 2   & bz = 16  & bz = 32  \\ \hline
MATIS Frame &  3.1 &  - & - & 13.11  & - & -   \\ 
MaskTrack-RCNN + SAM &  8.2 &  12.8 & - & 3.21 & 13.94 & -  \\ 
\textbf{SurgicalSAM} & \textbf{40.1}   & \textbf{57.4} & \textbf{59.8} & \textbf{1.92} & \textbf{5.90} & \textbf{9.56} \\
\hline
  & \multicolumn{6}{c}{Inference Speed (fps)} \\ \cline{2-7}  
 \multirow{-2}{*}{Method}  & \multicolumn{3}{c}{Online Feature Extraction} & \multicolumn{3}{c}{Offline Feature Extraction}\\ \hline
MaskTrack-RCNN + SAM &  \multicolumn{3}{c}{1.6} &  \multicolumn{3}{c}{14.3} \\
\textbf{SurgicalSAM} &  \multicolumn{3}{c}{\textbf{1.7}} & \multicolumn{3}{c}{\textbf{91.7} } \\
 \hline
\end{tabular}
\end{adjustbox}
\caption{Complexity Analysis.}
\label{tab:complex}
\end{table}

\subsection{Complexity Analysis}

To validate the efficiency of our method, we conduct a complexity analysis of SurgicalSAM against the best-performing zero-shot SAM baseline (MaskTrack-RCNN + SAM) and the SOTA specialist model MATIS Frame \cite{matis}. Their comparison regarding training efficiency across three batch sizes (bz) and inference efficiency is depicted in Table \ref{tab:complex}. In training, our method demonstrates considerably improved efficiency with notably faster speed and lower GPU memory consumption. Owing to the small number of tunable parameters, SurgicalSAM utilises less than 1/6 of the GPU memory of MATIS Frame with the same batch size, while achieving training over 10 times faster. In inference, the end-to-end pipeline of SurgicalSAM allows it to run faster than the complex multi-stage SAM baseline.

\section{Conclusion}
In this paper, we present SurgicalSAM, a novel method to efficiently tune SAM for surgical instrument segmentation. SurgicalSAM introduces a prototype-based class prompt encoder, which generates prompt embeddings directly from class prototypes. This eliminates the need for explicit points or bounding boxes from manual guidance or specialist detectors, enabling an end-to-end pipeline and enhancing prompt robustness.
We also introduce contrastive prototype learning to enhance the discriminative capability of class prototypes, improving differentiation among fine-grained instrument categories. 
Our method achieves state-of-the-art performance on both EndoVis2018 and EndoVis2017, demonstrating remarkable training and inference efficiency. It shows great promise for adapting SAM for surgical instrument segmentation.

\twocolumn[
  \begin{@twocolumnfalse}
  \begin{center}
           \LARGE
          \textbf{SurgicalSAM: Efficient Class Promptable Surgical Instrument Segmentation}
          \vspace{10pt}
 \\ Supplementary Material
  \vspace{10pt}
  \end{center}
  \end{@twocolumnfalse}
]

\section{Background and Motivation}

In this section, we offer additional background to clarify our motivation. We first highlight the clinical advantages of promptable segmentation over traditional methods in surgery. Then, we detail the two main problems with SAM that SurgicalSAM addresses: (1) SAM shows poor generalisation to surgical instruments due to the domain gap between surgical and natural objects, and (2) SAM requires a multi-stage pipeline with high prompting effort.

\subsection{Motivation of Promptable Segmentation}
Surgeon-computer interaction is essential in surgical practice. Ideally, surgeons should be able to provide reference information (\eg, an instrument category ID) for the system to then identify and segment the relevant targets. 
Yet, most existing methods for surgical instrument segmentation rely on conventional approaches that do not support such interactivity \cite{matis, baby}. 

To overcome this gap, in this work, we propose a promptable segmentation pipeline that inherently aligns with the needs of surgeons and offers an intuitive way for specifying the areas of interest. A promptable pipeline significantly enhances various components in surgical practice, including training \cite{surgical_training_survery23, planning}, planning \cite{planning}, navigation \cite{surgical_navigation_survery23}, and post-surgical analysis \cite{surgical_skill_survery22, surgical_data}. For instance, by incorporating a promptable pipeline within an augmented reality system, users can highlight and overlay specific areas on the surgical scene. This boosts engagement and interaction in surgical training and facilitates precise and customised navigation in robotic surgeries \cite{surgical_training_survery23, planning, surgical_navigation_survery23}.

\subsection{Motivation of SurgicalSAM}
We propose SurgicalSAM as a solution to two major problems in directly applying SAM to surgical instrument segmentation: (1) SAM's inferior zero-shot generalisation performance for surgical instruments due to their distinct characteristics from natural objects, and (2) the impracticality of SAM's requirement for accurate, per-frame explicit prompts (points or boxes) in surgical contexts.

\subsubsection{Surgical Instruments vs. Natural Objects}
SAM's zero-shot generalisation to surgical instruments is hindered by a lack of sufficient surgical data in its pre-training phase. This problem stems from the significant differences between surgical instruments and natural objects, which pose distinct challenges. Specifically, surgical instruments present a highly specialised appearance that is significantly different from natural objects (\eg, monopolar curved scissors vs. scissors). Also, surgical instruments operate on complex human tissues that differ from natural backgrounds. Thirdly, surgical instruments show high similarity across categories. 

Driven by the domain gap between surgical instruments and natural objects, we propose SurgicalSAM with prototype-based class prompt encoder to explicitly integrate surgical instrument knowledge with SAM's pre-trained knowledge through model tuning. Moreover, we propose contrastive prototype learning to improve the discrimination between prototypes and better differentiate fine-grained instrument classes. 

\subsubsection{Explicit Prompting of SAM}
To segment surgical instruments in a video, SAM requires explicit prompts in the form of points or bounding boxes at precise locations for each frame. This requires considerable human input or a well-performing detector for prompt preparation. Such a multi-stage pipeline with high prompting effort makes SAM impractical for direct use in surgical practice. 

In contrast, our method SurgicalSAM investigates surgical instrument segmentation prompted by category IDs, eliminating the need for explicit prompts, \ie, point-or-box prompts, and enabling a simpler, single-stage pipeline with greater prompting efficiency.
Moreover, our method eliminates the need for frame-by-frame prompt inputs, as a single class prompt can apply to multiple frames or an entire video. 

\section{Preliminary for Segment Anything Model}
We revisit the preliminaries of Segment Anything Model (SAM) in this section. SAM is a foundation model of image segmentation based on various visual prompts including points, bounding boxes, and masks \cite{sam}. It comprises three components, \ie, an image encoder, a prompt encoder, and a mask decoder, which are denoted as $E_I$, $E_P$, and $D_M$, respectively. 
Given an image $I$ and a prompt $P$, the image encoder employs Masked AutoEncoder (MAE) pre-trained Vision Transformer (ViT) \cite{vit} to derive image embedding $F_I$. Similarly, the prompt encoder converts the prompt into the corresponding prompt embedding $T_P$: 
\begin{equation}
F_I = E_I (I),
\end{equation}
\begin{equation}
T_P = E_P (P).
\end{equation}
Then, the mask decoder leverages the cross-attention mechanism to facilitate the interaction between the image embedding and the prompt embedding, ultimately resulting in a mask output:
\begin{equation}
M = D_M (F_I, [T_P, T_O]),
\end{equation}
where $T_O$ consists of learnable tokens that are concatenated with prompt embeddings to decode the final mask output $M$.

\section{Experiment Details and Results}

\subsection{Evaluation Metrics}
Challenge IoU measures the IoU between the predicted and ground-truth masks for only the classes present in an image, whereas IoU is computed across all classes. In our class promptable segmentation setting with class prompts provided, Challenge IoU and IoU yield identical results. We also report the IoU for each instrument class and compute their average as mean class IoU (mc IoU). Evaluations are conducted using the official code by \citet{isinet}. 

\subsection{Implementation Details}
The data of EndoVis18 \cite{endovis2018} and EndoVis17 \cite{endovis2017} are pre-processed to a resolution of 1024 $\times$ 1280, following the protocols defined by \citeauthor{ternausnet} \shortcite{ternausnet}. Employing established practices \cite{ternausnet, matis}, data augmentations are applied, including random flipping, random scale and crop, random rotation, and colour jitter.

During training, the class prototypes are initialised from the standard normal distribution $\mathcal{N}(0,1)$ and gradually updated by the loss functions. 

\subsection{Baseline Implementation Details}

We provide details for implementing the detection-based, tracking-based, and reference-based zero-shot SAM baselines for surgical instrument segmentation. 

\subsubsection{Detection-based Zero-Shot SAM Baseline}
Detection-based zero-shot SAM frameworks consist of two stages: (1) employing a trained detector to predict bounding boxes from an input image; and (2) feeding the predicted bounding boxes as prompts into SAM for mask prediction. To ensure a fair comparison, we implement this baseline in a class promptable manner, ensuring a consistent setting with our method. Specifically, the detector predicts a set of candidate bounding boxes $\{\mathcal{B}_k\}_{k \in \{1,2,...,C\}}$, each associated with a predicted class label and a predicted confidence score, where $\mathcal{B}_k$ represents the predicted bounding boxes for class $k$. Then, given class $c$ as a prompt, only the bounding boxes corresponding to class $c$, \ie, $\mathcal{B}_c$, are considered. If $\mathcal{B}_c$ contains a single bounding box, it will be fed into SAM for mask prediction. Otherwise, if $\mathcal{B}_c$ contains multiple bounding boxes, then a threshold $\alpha$ is used to filter $\mathcal{B}_c$, and only those bounding boxes with confidence scores greater than $\alpha$ will be fed into SAM for mask prediction. The masks obtained from all the filtered bounding boxes are combined as the final mask for class $c$. 

We consider two popular detector backbones as our detection-based zero-shot SAM baselines, i.e., MaskTrack-RCNN \cite{masktrackrcnn} and Mask2Former\cite{m2f}, respectively. 
For MaskTrack-RCNN, we manually provide the video-level instance labels for the EndoVis2018 \cite{endovis2018} dataset and train MaskTrack-RCNN with the detector module pre-trained on COCO \cite{mask_rcnn,coco}. A batch size of 8 and an SGD optimiser with a learning rate of 0.005 are used in training. For Mask2Former, we directly use the trained model provided by \citet{matis} for both EndoVis2018 \cite{endovis2018} and Endovis2017 \cite{endovis2017} datasets. The threshold $\alpha$ is set to 0.7 empirically for MaskTrack-RCNN and 0.5 for Mask2Former following \citet{matis}.

To measure the inference speed of this multi-stage pipeline, we separately capture the time for stage (1) acquiring candidate bounding boxes from all frames using the trained detector, and stage (2) feeding all filtered boxes into SAM to obtain masks. We denote the inference time for the two stages as $\mathcal{T}_1$ and $\mathcal{T}_2$, respectively. The final inference speed is computed as $Q/(\mathcal{T}_1 + \mathcal{T}_2)$ in fps (frames per second), where $Q$ refers to the total number of frames. 

\subsubsection{Tracking-based Zero-Shot SAM Baseline}
Regarding the tracking-based zero-shot SAM baseline, we employ the official implementation of Track Anything \cite{trackanything}. For a given video, we provide ground truth-based point prompts for all the instruments in the frames where new instances appear. In other words, at first, point prompts are provided for all instrument instances in the first frame for tracking initialisation. Then, when a new instance appears at frame $t$, we provide point prompts for all the instrument instances in frame $t$ and then the tracking continues. The above process is repeated until the end of the video is reached. 

\begin{figure}[!t]
    \centering
    \subfloat[One-Point Prompting]{%
      \includegraphics[width=0.23 \textwidth]{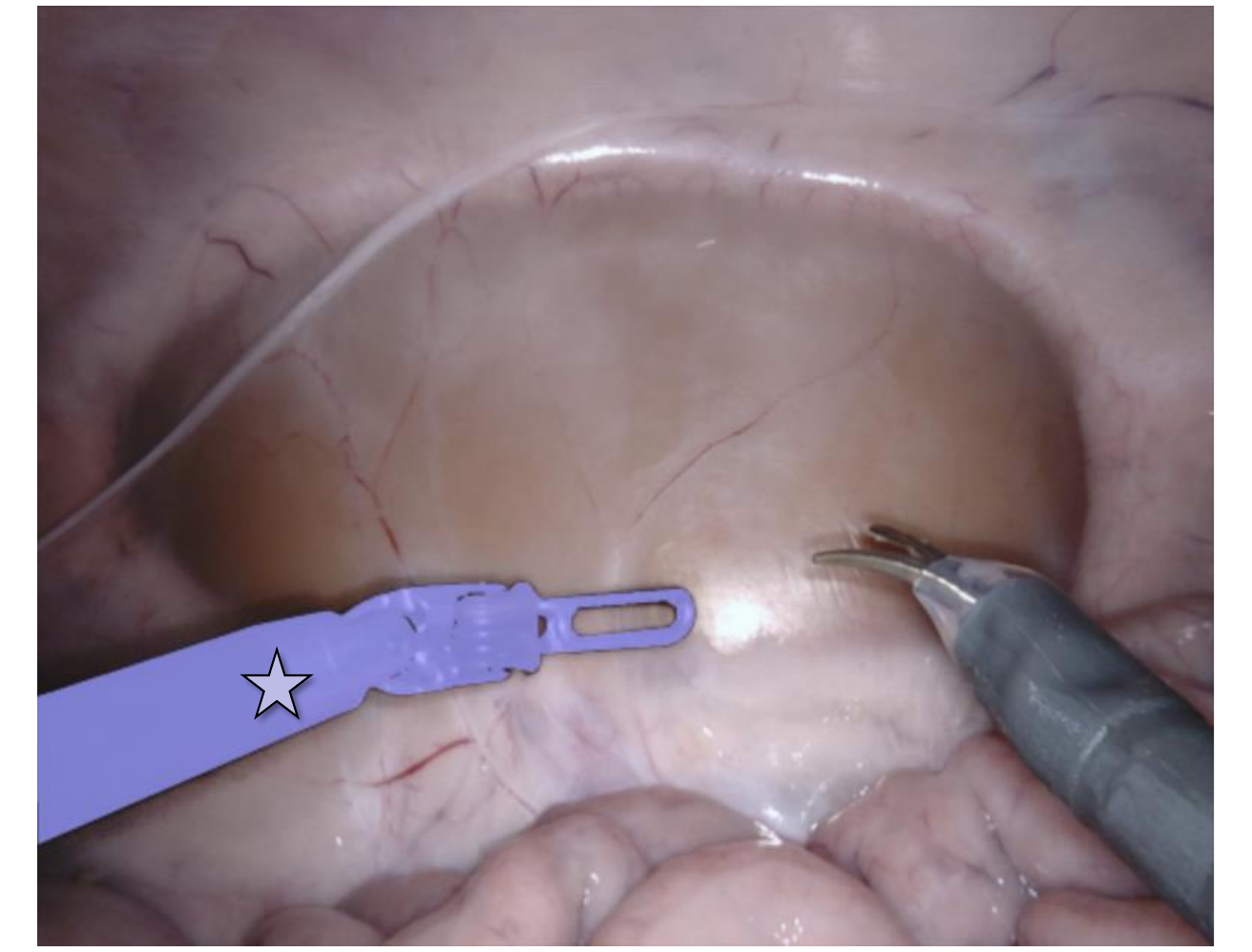}
    }
    \subfloat[Five-Point Prompting]{%
      \includegraphics[width=0.23 \textwidth]{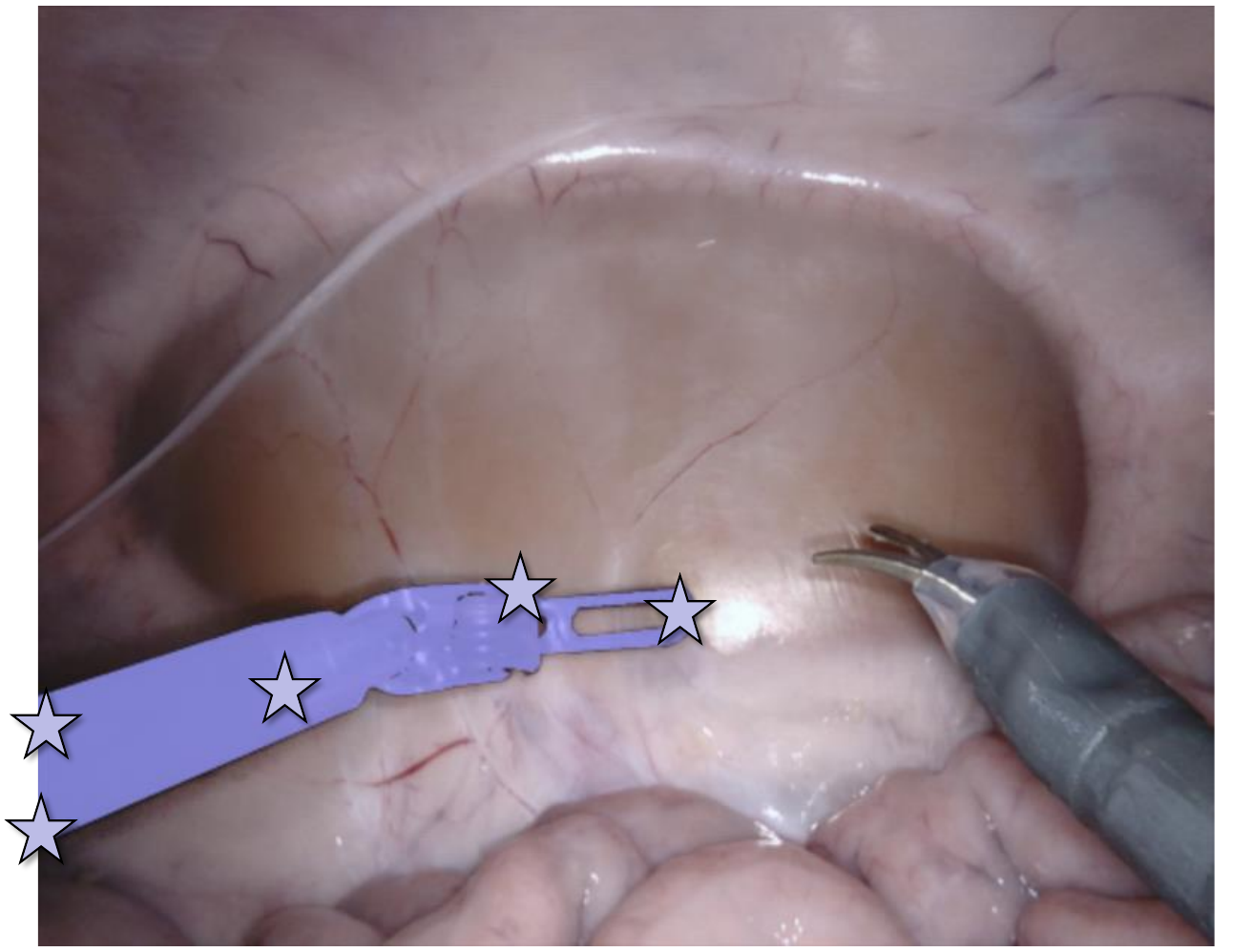}
    }
    \caption{Illustration of using one point and five points to prompt Track Anything \cite{trackanything}. (a) The centroid is used as a single-point prompt. (b) The centroid and four extremity points (leftmost, rightmost, topmost, and bottommost points, each with a margin of 10 pixels from the boundary) are used as the prompts.}
   \label{fig:point_mode}
 \end{figure}

We implement two types of point prompting: the one-point and the five-point prompting methods. In the one-point prompting method, the centroid of the ground-truth mask is used as the prompt. In the five-point prompting method, the centroid and four extremity points (leftmost, rightmost, topmost, and bottommost points, each with a margin of 10 pixels from the boundary) of the ground-truth mask are utilised to prompt Track Anything. We present an illustration of both prompting methods in Fig. \ref{fig:point_mode}. 

\subsubsection{Reference-based Zero-Shot SAM Baseline}
We employ the official implementation of PerSAM \cite{persam} for our reference-based zero-shot SAM baseline. To predict the mask of class $c$ in a given frame, we utilise the frame where class $c$ first appears in the video as the reference frame and the corresponding ground-truth mask as the reference mask for PerSAM. Following \citet{persam}, we explore both the training-free and the fine-tuning versions of PerSAM, and use the default settings for both model configuration and training setup. 

\subsection{Ablation Study of Prompt Embeddings}
We conduct an ablation study of the prompt embeddings of SurgicalSAM on EndoVis2018 \cite{endovis2018}. Specifically, we remove the dense prompt embeddings, sparse prompt embeddings, and the positive and negative embeddings from the prototype-based class prompt encoder. The results, as shown in Table \ref{tab:ablat_addition}, demonstrate the effective role of each component. Particularly, removing dense prompt embeddings leads to a more significant performance drop than removing sparse ones, confirming our expectation that dense prompt embeddings, functioning as masks in SAM, hold more information and guide more precise decoding than the point-based sparse prompt embeddings. Our proposed SurgicalSAM effectively utilises both types of prompt embeddings to achieve optimal results.

\begin{table}[htbp] 
\renewcommand{\arraystretch}{1}
\setlength{\tabcolsep}{15pt}
\centering
\begin{adjustbox}{width=0.45\textwidth}
\begin{tabular}{l c c }
\hline
  Method& Challenge IoU& mc IoU\\
\hline
\textbf{SurgicalSAM (Ours)} & \textbf{80.33}& \textbf{58.87}     \\
 w/o Dense Embed.& 23.61&14.13 \\
 w/o Sparse Embed. & 78.58&58.24 \\
 w/o Pos\&Neg Embed.& 77.69& 57.47 \\
 \hline
\end{tabular}
\end{adjustbox}
\caption{Ablation Study on Prompt Embeddings.}
\label{tab:ablat_addition}
\end{table}

\subsection{Comparison with Text Promptable Baseline}
To provide an additional ablation study of the proposed contrastive prototype learning, we build a text promptable baseline. This baseline uses the CLIP \cite{clip} text embeddings of the class names as the prototypes, without training with the prototype contrastive loss $\mathcal{L}_{PCL}$. It achieves 75.94\% for Challenge IoU and 51.76\% for mc IoU, showing the superiority of our method.

\subsection{Complexity Analysis and Clinical Significance}
SurgicalSAM contains 641.68M parameters in total, with 637.03M parameters from a frozen ViT-H image encoder and only 4.65M parameters for tuning. As shown in Table \ref{tab:complex}, SurgicalSAM has very low training and inference costs. High training efficiency is crucial for real-world surgical applications. For example, it improves the resource efficiency during model development and makes surgical technology more affordable to healthcare institutions and more adaptable to specific surgical procedures (e.g., via efficient model tuning).

\subsection{Visualisation}
\subsubsection{Result Visualisation}
\begin{figure*}[h]
\centering
\includegraphics[width=0.95\textwidth]{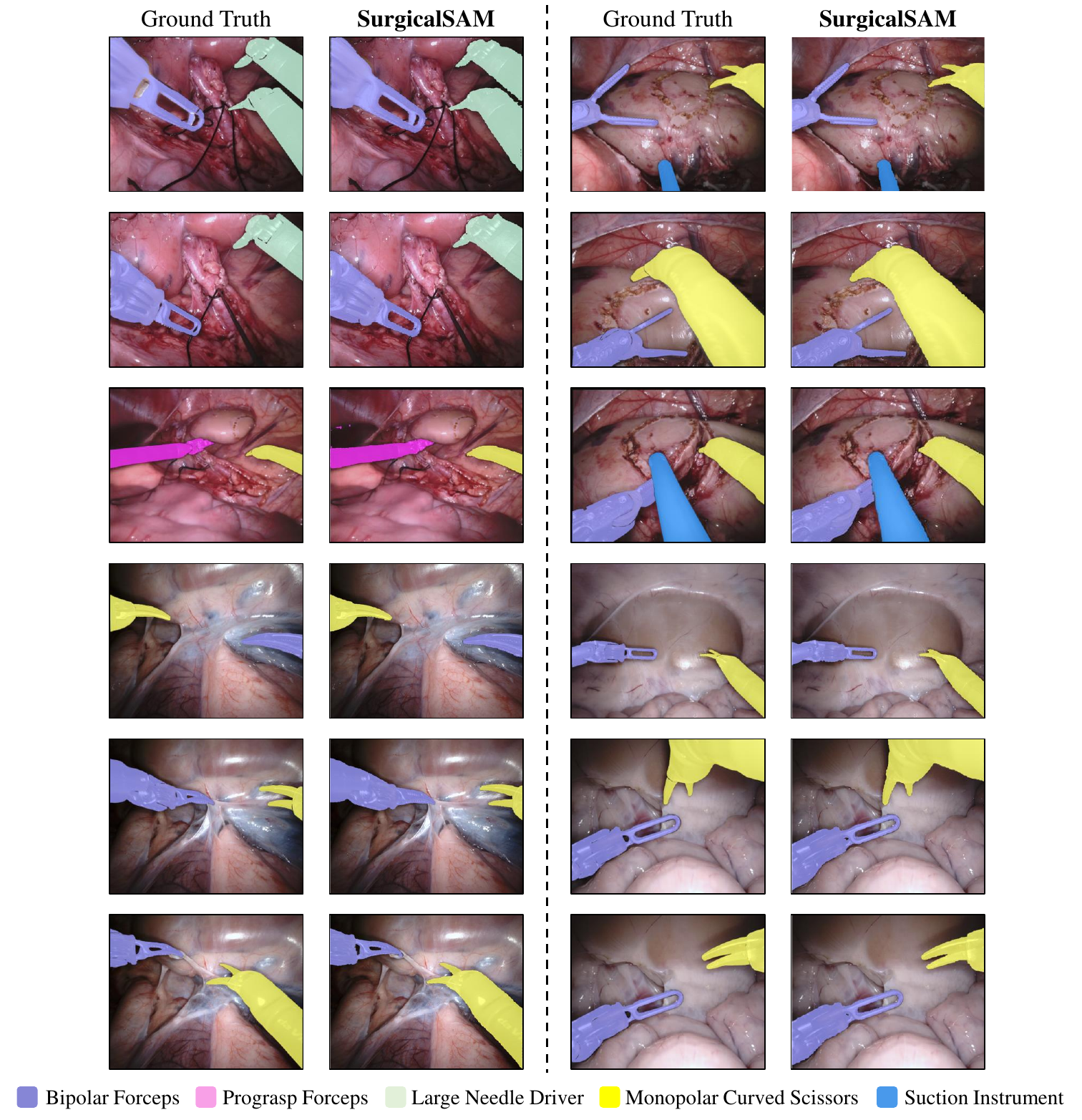}
\caption{Visualisation of Predicted Masks on the EndoVis2018 Dataset~\cite{endovis2018}.} 
\label{fig:more_results_endovis18}
\end{figure*}

\begin{figure*}[h]
\centering
\includegraphics[width=0.95\textwidth]{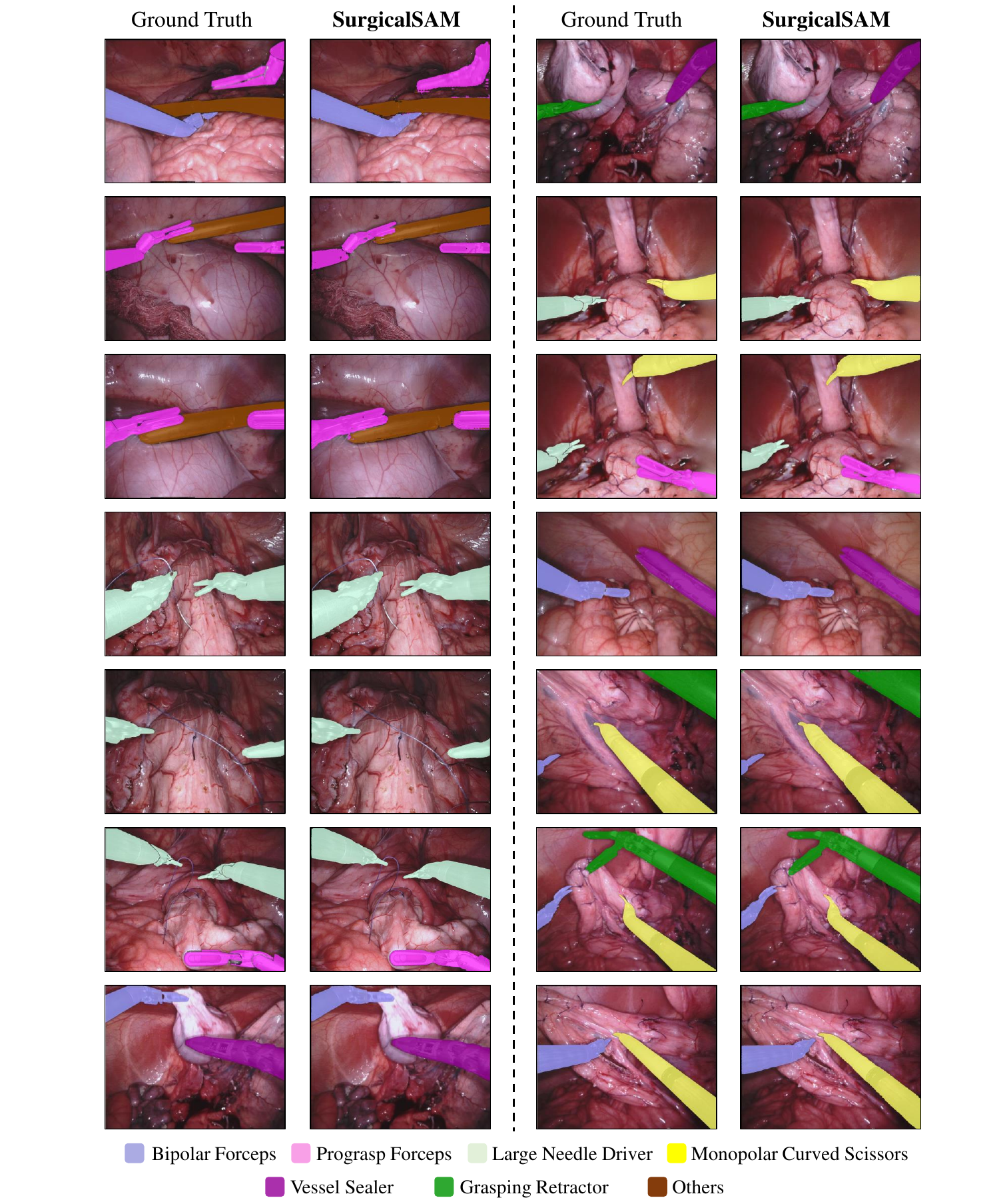}
\caption{Visualisation of Predicted Masks on the EndoVis2017 Dataset~\cite{endovis2017}.} 
\label{fig:more_results_endovis17}
\end{figure*}

We present additional visualisation results of the predictions by SurgicalSAM on EndoVis2018 \cite{endovis2018} and EndoVis2017 \cite{endovis2017} datasets in Fig. \ref{fig:more_results_endovis18} and Fig. \ref{fig:more_results_endovis17}, respectively. Samples from all videos in each dataset are included to comprehensively showcase the promising performance of our method. Notably, our method produces high-quality masks that precisely delineate instrument boundaries. Furthermore, in cases where two or three categories are present in an image, SurgicalSAM adeptly distinguishes between them and accurately identifies the prompted category, benefiting from the discriminative prototypes learned through contrastive prototype learning. 

\subsubsection{Similarity Map Visualisation}
\begin{figure*}[h]
\centering
\includegraphics[width=0.9\textwidth]{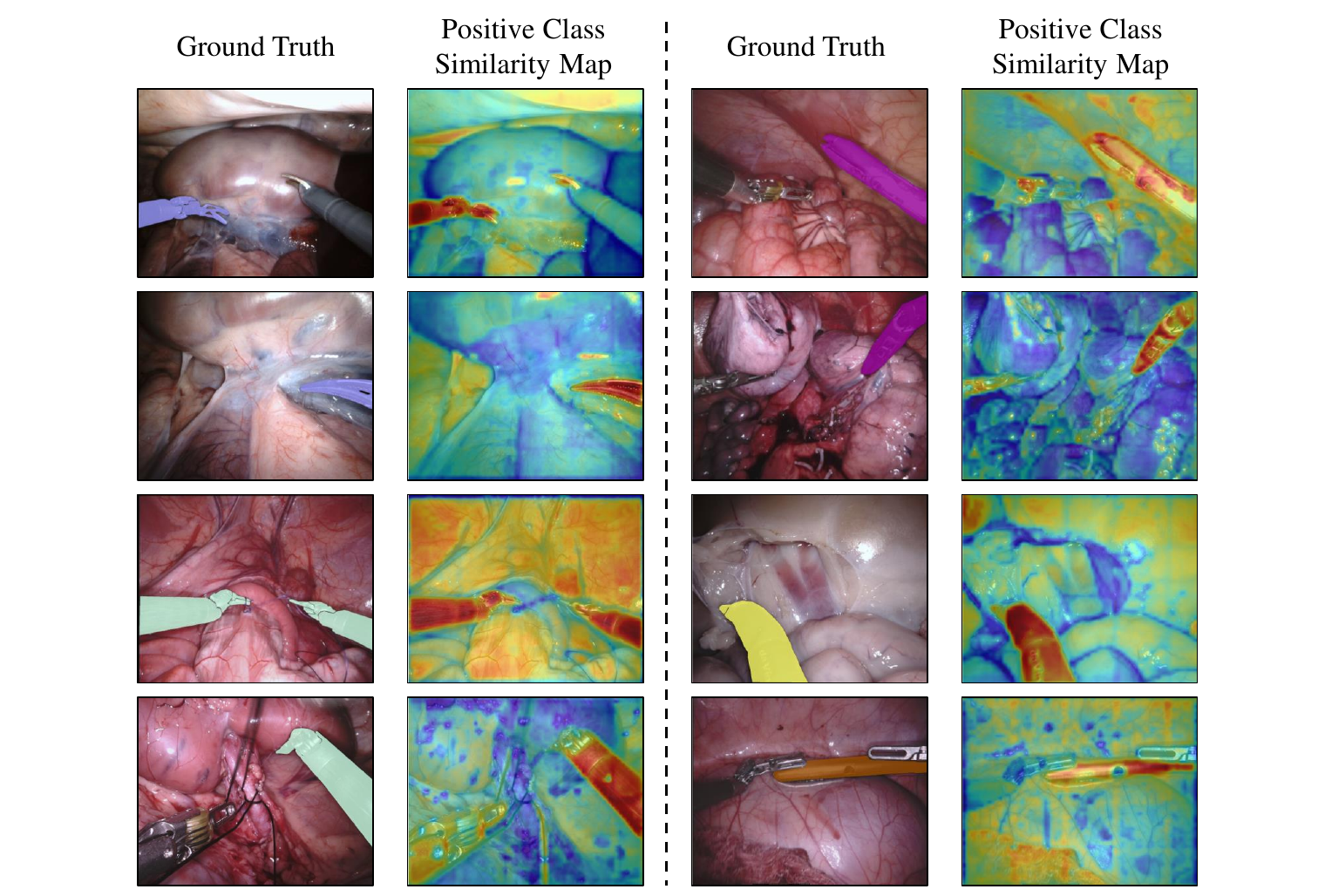}
\caption{Visualisation of Positive Class Similarity Maps.} 
\label{fig:att_vis}
\end{figure*}

To better validate the class activation mechanism of our proposed prototype-based class prompt encoder, in Fig. \ref{fig:att_vis} we provide visualisation results of the positive class similarity maps $S^{(c)}$, computed between the image embedding and the prototype of the prompted class $c$. $S^{(c)}$ is normalised for visualisation purpose and the ground-truth mask of the prompted class is also presented in Fig. \ref{fig:att_vis} for reference. 
The visualisation results validate the capability of SurgicalSAM to accurately identify class-specific regions for generating relevant class-activated features that enables accurate recognition and localisation of various fine-grained instrument categories.

\subsection{Result Reproducibility}
For reproducibility, the results presented in our main paper are obtained using a seed value of 666. We further conduct additional experiments with different seed values on EndoVis2018 with results shown in Table \ref{tab:results_repeat}. Consistent results across different repeats can be observed, with a mean of 80.16 and a standard deviation of 0.29 for Challenge IoU and a mean of 60.20 and a standard deviation of 1.42 for mc IoU, confirming robustness against the randomness w.r.t. random seeds. 

\begin{table*}[h]
\centering
\renewcommand{\arraystretch}{1}
\setlength{\tabcolsep}{12pt}
\begin{adjustbox}{width=0.97\textwidth}
\begin{tabular}{ c c c c c c c c c c}
\hline
   &  &  & \multicolumn{7}{c}{Instrument Categories}\\ \cline{4-10} 
  \multirow{-2}{*}{Challenge IoU} & \multirow{-2}{*}{IoU} & \multirow{-2}{*}{mc IoU} & BF & PF & LND & SI & CA & MCS & UP \\ \hline
\textbf{80.33} & \textbf{80.33} &\textbf{ 58.87} & \textbf{83.66} & \textbf{65.63} & \textbf{58.75}  & \textbf{54.48} & \textbf{39.78} & \textbf{88.56} & \textbf{21.23}  \\
79.75 & 79.75 &  59.57 & 83.78 & 63.44 & 50.40 & 48.15 & 49.01 & 89.28 & 32.94  \\
80.41 & 80.41 & 62.16 & 84.01 & 68.31 & 58.48 & 54.81 & 50.59 & 88.7 & 30.25 \\
\hline
\end{tabular}
\end{adjustbox}
\caption{Repeated Experiments on EndoVis2018 Dataset. The results reported in the main paper are shown in \textbf{bold}.}
\label{tab:results_repeat}
\end{table*}

\section{Enlarged Figures}
\begin{figure*}[!t]
\centering
\includegraphics[width=0.8\textwidth]{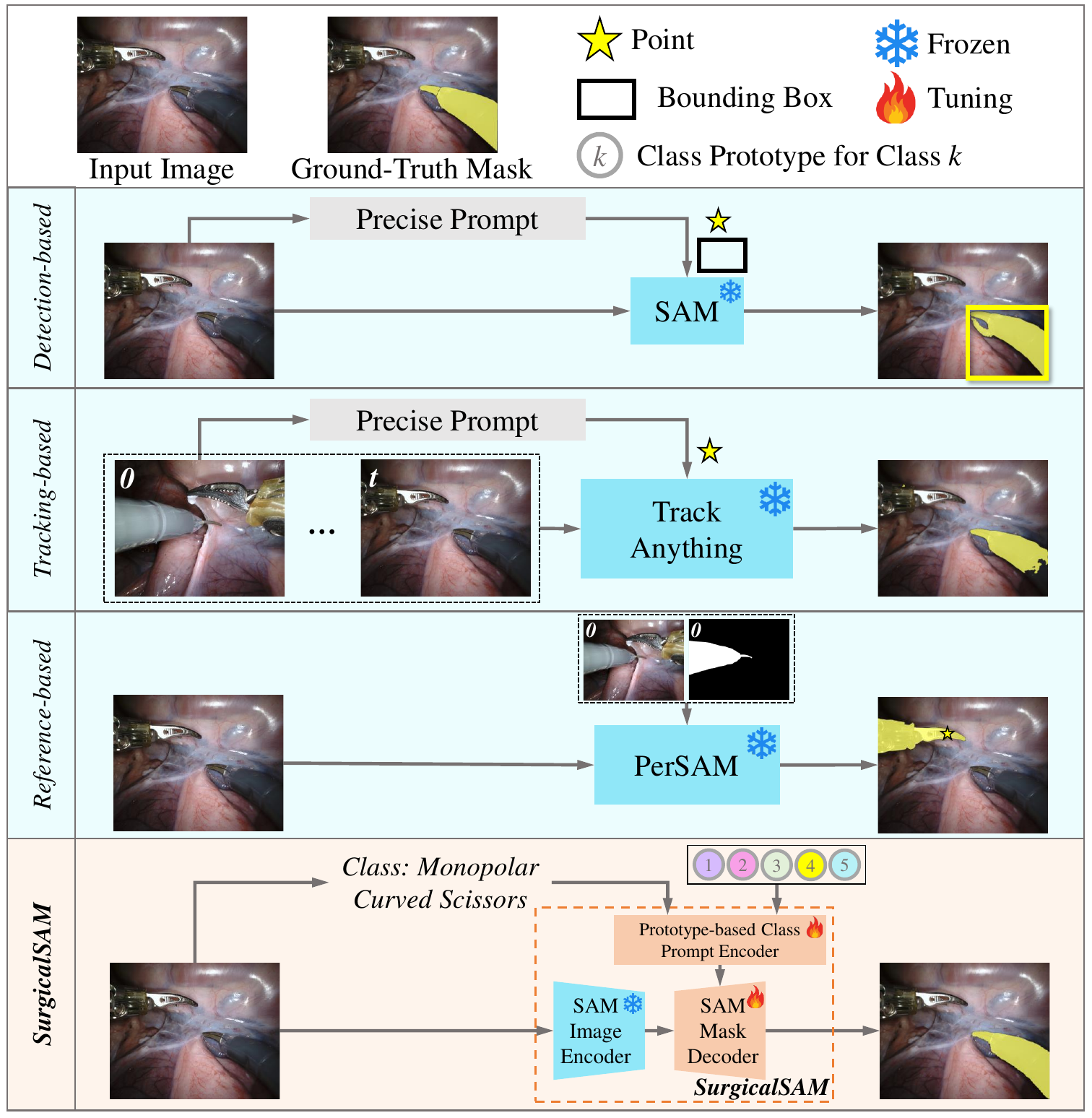}
\caption{Enlarged Version of Fig. 1 in the Main Paper - Comparison of our SurgicalSAM against existing detection-based, tracking-based, and reference-based zero-shot SAM frameworks for surgical instrument segmentation. } 
\label{fig:enlarged_motiv_compare}
\end{figure*}

\begin{figure*}[!t]
\centering
\includegraphics[width=0.6\textwidth]{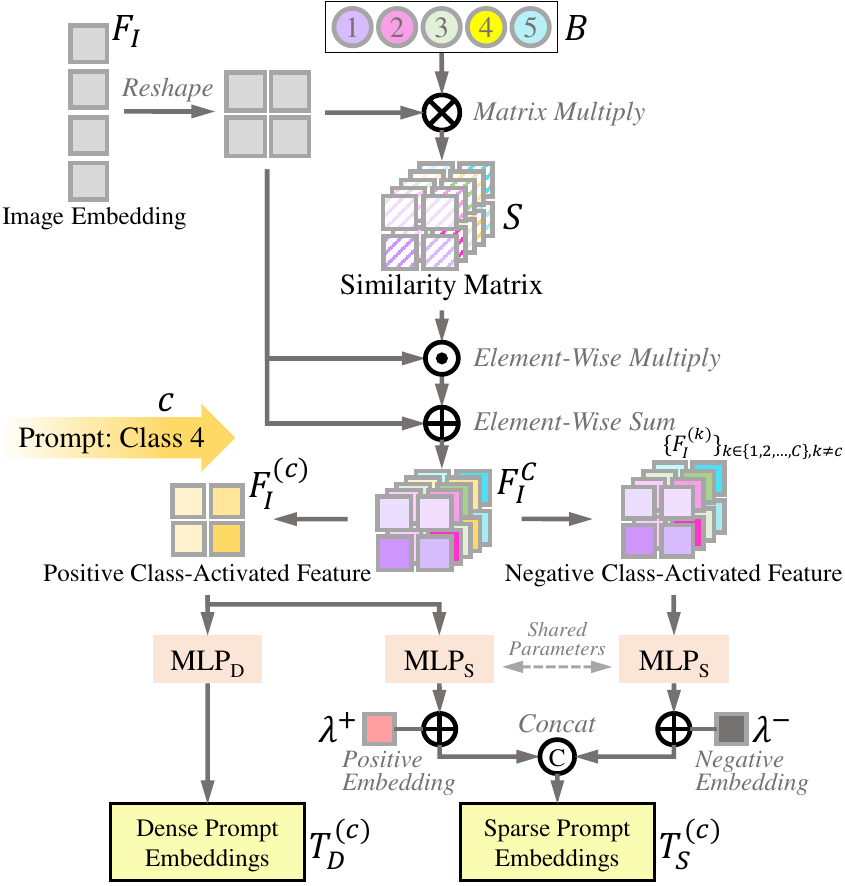}
\caption{Enlarged Version of Fig. 3(b) in the Main Paper - Prototype-based Class Prompt Encoder. } 
\label{fig:enlarged_pcpe}
\end{figure*}

\begin{figure*}[!t]
\centering
\includegraphics[width=0.8\textwidth]{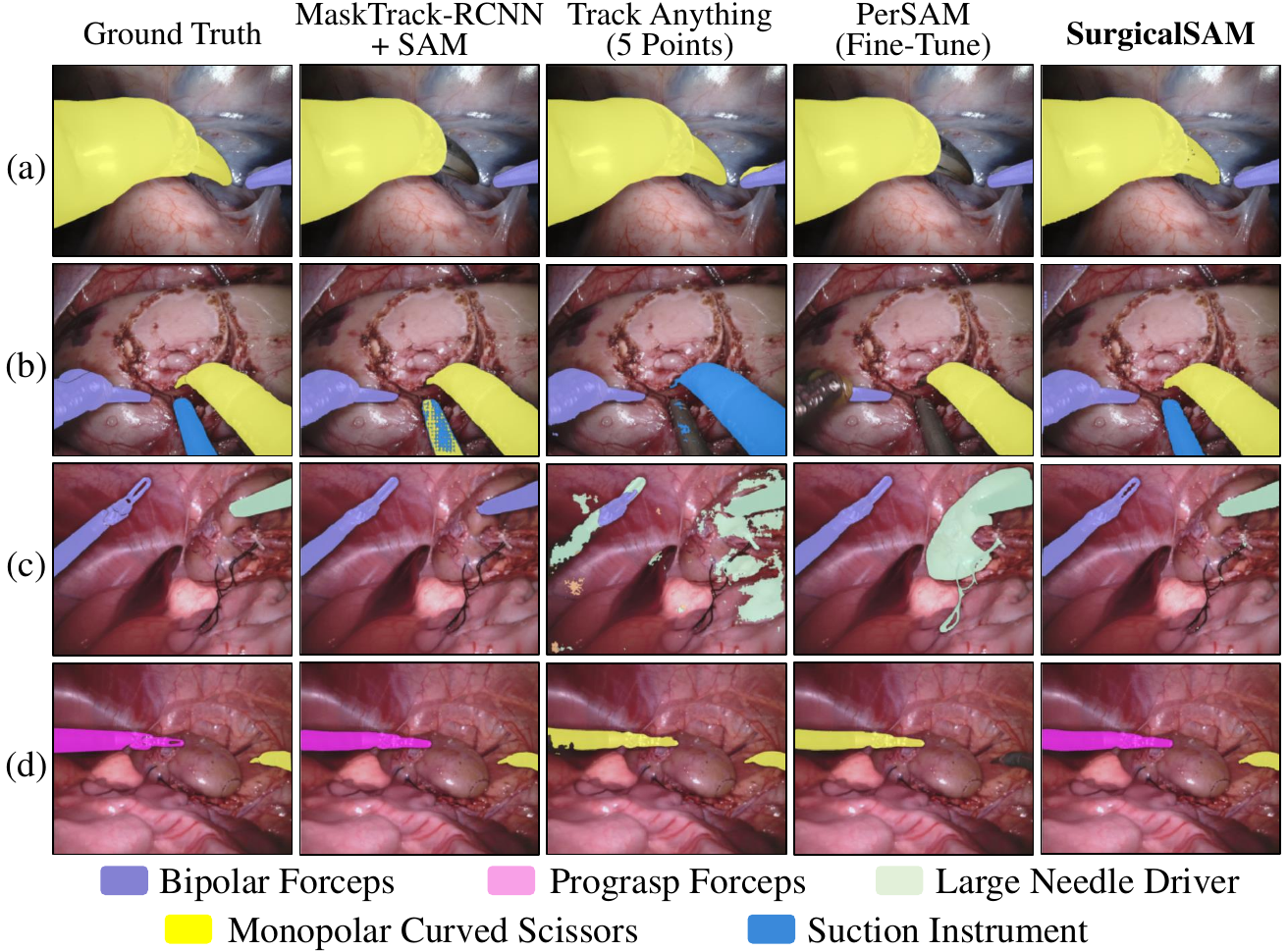}
\caption{Enlarged Version of Fig. 5 in the Main Paper - Visualisation of Predicted Masks.} 
\label{fig:enlarged_vis_results}
\end{figure*}

Fig. \ref{fig:motiv_compare}, Fig. \ref{fig:method_overview}(b), and Fig. \ref{fig:vis_results} in the main paper are best to be viewed on screen with zoom-in for better clarity. To facilitate better visibility and convenience, we offer enlarged versions of these figures here in Fig. \ref{fig:enlarged_motiv_compare}, Fig. \ref{fig:enlarged_pcpe}, and Fig. \ref{fig:enlarged_vis_results}. 

\section{Acknowledgements}
This study was partially supported by Australian Research Council (ARC) grant DP210102674.

\bibliography{aaai24}
\end{document}